\documentclass[final]{siamltex}
\usepackage{verbatim}
\usepackage{graphicx, epsfig}
\usepackage{float}
\usepackage{amsmath,amssymb}
\usepackage{hyperref}
\usepackage{fancyref}
\usepackage{xcolor}
\usepackage{subfig}
\usepackage{multirow}
\usepackage{bbm}
\usepackage{float}
\usepackage{soul}
\usepackage[makeroom]{cancel}
\usepackage[normalem]{ulem}
\usepackage{bm}
\usepackage{tikz}
\usepackage{algorithm}
\usepackage{algpseudocode}
\usepackage{comment}
\newtheorem{remark}{Remark}[section]

\title{Efficient Bayesian Physics Informed Neural Networks for inverse problems via Ensemble Kalman Inversion}
\author{Andrew Pensoneault\thanks{Department of Mathematics, University of Iowa, Iowa City, IA 52246, USA. Email: 
andrew-pensoneault@uiowa.edu.}
\and Xueyu Zhu\thanks{Department of Mathematics, University of Iowa, Iowa City, IA 52246. USA. Email: xueyu-zhu@uiowa.edu.}
}

\begin{document}
\maketitle

\begin{abstract}
Bayesian Physics Informed Neural Networks (B-PINNs) have gained significant attention for inferring physical parameters and learning the forward solutions for problems based on partial differential equations. However, the overparameterized nature of neural networks poses a computational challenge for high-dimensional posterior inference. Existing inference approaches, such as particle-based or variance inference methods, are either computationally expensive for high-dimensional posterior inference or provide unsatisfactory uncertainty estimates. 
In this paper, we present a new efficient inference algorithm for B-PINNs that uses Ensemble Kalman Inversion (EKI) for high-dimensional inference tasks. By reframing the setup of B-PINNs as a traditional Bayesian inverse problem, we can take advantage of EKI's key features: (1) gradient-free, (2) computational complexity scales linearly with the dimension of the parameter spaces, and (3)  rapid convergence with typically $\mathcal{O}(100)$ iterations.  We demonstrate the applicability and performance of the proposed method through various types of numerical examples. We find that our proposed method can achieve inference results with informative uncertainty estimates comparable to Hamiltonian Monte Carlo (HMC)-based B-PINNs with a much reduced computational cost. These findings suggest that our proposed approach has great potential for uncertainty quantification in physics-informed machine learning for practical applications.
\end{abstract}


\pagestyle{myheadings}
\thispagestyle{plain}

\section{Introduction}
\label{sec:Intro}
Many applications in science and engineering can be accurately modeled by partial differential equations (PDEs). These equations often contain parameters corresponding to the physical properties of the system. In practice, these properties are often challenging to measure directly. Inverse problems arise when indirect measurements of the system are used to infer these model parameters. These problems are often ill-posed, with the available measurements being limited and noisy. Traditional approaches for solving these problems can be sensitive to noise and data scarcity. In addition, they often require many runs of sophisticated forward numerical solvers \cite{Willard_2020}, which can be computationally expensive. Furthermore, it is not uncommon that the initial or boundary conditions are missing for real-world applications. In such a case, the traditional numerical forward solver might not even be able to run. To address these issues, there is growing interest in developing more efficient, robust, and flexible alternatives.

Recently, Scientific Machine Learning (Sci-ML) \cite{baker_2019, RAISSI2019, Lu_2021, Psaros_2022}, a set of approaches that combine domain-specific scientific and engineering knowledge with powerful machine learning tools, has received much attention. These approaches have proven effective in solving PDE-based inverse problems efficiently. One particularly promising approach is Physics Informed Neural Networks (PINNs) \cite{RAISSI2019, karniadakis2021physics}. PINNs construct a neural network approximation of the forward solution to the underlying problem while inferring model parameters simultaneously by minimizing data misfit regularized by the underlying governing PDE residual loss. In addition to obtaining estimates of these quantities, the presence of noise and lack of data make it essential to quantify the impact of uncertainty on the surrogate and parameters, especially in high-stakes applications \cite{zou2022neuraluq, Psaros_2022}. However, standard PINN approaches provide only deterministic estimates. Non-Bayesian approaches to uncertainty quantification (UQ) in PINN, such as neural network dropout \cite{Yang_2021}, have been explored. Despite their efficiency, the estimates provided by these methods tend to be less satisfactory, as pointed out in \cite {Yang_2021, Jiang2022}.

Alternatively, several attempts to develop Bayesian PINNs (B-PINNs) have been explored, enabling uncertainty quantification in the neural network approximations and the corresponding physical parameter estimates. These methods utilize Bayesian Neural Networks (BNNs) as surrogates by treating weights of neural networks and physical parameters as random variables. In general, Markov Chain Monte Carlo (MCMC) methods are one of the most popular approaches for Bayesian inference tasks. These methods approximate the posterior distribution of parameters with a finite set of samples. While these approaches provide inference with asymptotic convergence guarantees, the use of standard MCMC methods in deep learning applications is limited by their poor scalability for large neural network architectures and large data sets \cite{papamarkou2022challenges}. In practice, Hamiltonian Monte Carlo (HMC) is the gold standard for inference for Bayesian neural networks \cite{cobb_2020}. HMC is an MCMC method that constructs and solves a Hamiltonian system from the posterior distribution. This approach enables higher acceptance rates than traditional MCMC methods. In the context of B-PINNs \cite{Yang_2021}, an HMC-based B-PINN has been investigated for inference tasks. Despite this, the computational cost of inference remains high.

Alternatively, Variational Inference (VI) approaches to B-PINNs \cite{Yang_2021} posit that the posterior distribution of the physical and neural network parameters lives within a family of parameterized distributions and solves a deterministic optimization problem to find an optimal distribution within that family \cite{Blei_2017}. In general, VI approaches are more efficient than HMC approaches and scale well for large parameter spaces and data sizes. Variational inference methods, however, do not share the same theoretical guarantees as MCMC approaches and only provide estimates within the function space of the family of parameterized distributions. Additionally, the inference quality depends on the space of parameterized densities chosen. For example, in \cite{Yang_2021}, KL divergence-based B-PINNs tend to provide less satisfactory estimates than the corresponding HMC B-PINNs and are less robust to measurement noise.

The particle-based VI approach bridges the gap between VI and MCMC methods, combining the strengths of both approaches to provide more efficient inference than MCMC methods while also providing greater flexibility through non-parametric estimates in contrast with the parametric approximations utilized in VI \cite{yan_2021}. In the context of B-PINNs, Stein Variational Gradient Descent (SVGD) has been proposed for inference tasks to reconstruct idealized vascular flows with sparse and noisy velocity data \cite{Yang_2021}. However, SVGD tends to underestimate the uncertainty of the distribution for high dimensional problems, collapsing to several modes \cite{ba2021understanding, Wang2017}.

Recently, Ensemble Kalman inversion (EKI) \cite{Iglesias_2013, Kovachki_2019, Iglesias_2016} has been introduced as a particle-based VI approach that uses the Ensemble Kalman filter algorithm to solve traditional Bayesian inverse problems. EKI methods have many appealing properties: they are gradient-free, easily parallelizable, robust to noise, and computationally scale linearly with ensemble size \cite{LopezGomez_2022}. Additionally, these methods use low-rank approximate Hessian (and gradient) information, allowing for rapid convergence of the methods \cite{Zhang_2022}. In the case of a Gaussian prior and linear Gaussian likelihood \cite{Gland2009}, Ensemble Kalman methods have asymptotic convergence to the correct posterior distribution in a Bayesian sense. Ensemble Kalman methods are asymptotically biased when these assumptions are violated, yet they are computationally efficient compared to asymptotically unbiased methods and empirically provide reasonable estimates \cite{DUFFIELD2022109523, Botha2022}. Recently, these methods have also been used to train neural networks efficiently \cite{Kovachki_2019, haber2018never,chen2022novel, Guthweissman_2020EnsembleKF, Zhang_2022}. In \cite{Kovachki_2019},  EKI was first proposed as a derivative-free approach to train the neural networks but primarily for traditional purely data-driven machine learning problems. Recently, a one-shot variant of EKI \cite{Guthweissman_2020EnsembleKF} has been used to learn maximum a posteriori (MAP) estimates of a NN surrogate and model parameters for several inverse problems involving PDEs; however, this approach still requires traditional numerical solvers or discretizations to enforce the underlying physical laws, which can be computationally expensive for large scale complex applications. Additionally, while EKI has been traditionally used to obtain MAP estimates of unknown parameters, recent works have begun to investigate the use of the EKI methods to efficiently draw approximate samples for Bayesian inference for traditional Bayesian inverse problems \cite{Botha2022, DUFFIELD2022109523,huang2022, huang2022iterated,chen2012ensemble}.

Motivated by the recent advances in EKI, we present a novel efficient inference method for B-PINNs based on EKI. Specifically, we first recast the setup of B-PINNs as a traditional Bayesian inverse problem so that EKI can be applied to obtain approximate posterior estimates. Furthermore, based on the variant of EKI in \cite{Kovachki_2019}, we present an efficient sampling-based inference approach to B-PINNs. Because our approach inherits the properties of EKI, it provides efficient gradient-free inference and is well suited for large-scale neural networks thanks to the linear computational complexity of the dimension of unknown parameters \cite{LopezGomez_2022}. Further, unlike the traditional setting for EKI \cite{Iglesias_2013}, this approach replaces the expensive numerical forward solver with a NN surrogate trained on the measurements and the underlying physics laws to jointly estimate the physical parameters and forward solution with the corresponding uncertainty estimation. Because the trained neural network surrogate is cheap to evaluate, our proposed approach can draw a large number of samples efficiently and is thus expected to reduce sampling errors significantly. Through various classes of numerical examples, we show that the EKI method can efficiently infer physical parameters and learn valuable uncertainty estimates in the context of B-PINNs.  Furthermore, the empirical results show that EKI B-PINNs can deliver comparable inference results to HMC B-PINNs while requiring significantly less computational time. To our best knowledge, this is the {\it first attempt} to use EKI for efficient inference under the context of B-PINNs. 

The rest of the paper is organized as follows: in section \ref{sec:bg}, we first introduce the setup of the PDE-based inverse problem and briefly review B-PINNs. In section \ref{sec:EKI-section}, we first reframe the problem in terms of a Bayesian Inverse problem and introduce EKI under the context of Bayesian inverse problems. We then introduce the building blocks of our proposed EKI B-PINNs framework. We demonstrate the performance of this approach via several numerical examples in section \ref{sec:NumExample}. Finally, we conclude in section \ref{sec:summary}.
\section{Problem Setup and Background}
\label{sec:bg}

\label{subsec:problem}
We consider the following partial differential equation (PDE): 
\begin{align}
\mathcal{N}_x(u(\mathbf{x});\boldsymbol{\lambda})&=f(\mathbf{x}) \quad \mathbf{x}\in \Omega, \label{al:pde}\\
\mathcal{B}_x(u(\mathbf{x});\boldsymbol{\lambda})&=b(\mathbf{x}) \quad \mathbf{x}\in \partial\Omega,
\end{align}
where $\mathcal{N}_x$ and $\mathcal{B}_x$ denote the differential and boundary operators, respectively. The spatial domain $D\subseteq\mathbb{R}^d$ has boundary $\Gamma$, and $\boldsymbol{\lambda}\in\mathbb{R}^{N_\lambda}$ represents a vector of unknown physical parameters. The forcing function $f(\mathbf{x})$ and boundary function $b(\mathbf{x})$ are given, and $\mathbf{u}(\mathbf{x})$ is the solution of the PDE. For time-dependent problems, we consider time $t$ as a component of $\mathbf{x}$ and consider domain $\Omega$ and boundary $\partial\Omega$ to additionally contain the temporal domain and initial boundary, respectively.

In this setting, we additionally have access to $N_u$ measurements $\mathcal{D}_u=\{(\mathbf{x}_u^i, u^i)\}_{i=1}^{N_u}$ of the forward solution at various locations. Given the available information, we aim to infer the physical parameters $\boldsymbol{\lambda}$ with uncertainty estimation.

\subsection{Bayesian Physics Informed Neural Networks (B-PINNs)}
\label{subsec:B-PINN} 

Over the last several years, SciML approaches for solving inverse problems have received much attention \cite {baker_2019,karniadakis2021physics}. One of the most promising approaches is the Physics Informed Neural Networks (PINNs), which approximate the forward solution $u(\mathbf{x})$ with a fully connected neural network surrogate $\tilde{u}(\mathbf{x};\boldsymbol{\theta})$, parameterized by neural network's weight parameter $\boldsymbol{\theta}\in\mathbb{R}^{N_\theta}$. Denote $\boldsymbol{\xi}=[\boldsymbol{\theta},\boldsymbol{\lambda}]$ as the concatenation of the neural network and physical parameters. The neural networks are trained by minimizing a weighted sum of the data misfit regularized by initial/boundary data misfit and underlying PDE residual loss \eqref{al:pde} over a set of discrete points  - 
 ``residual points" and ``boundary points," respectively, in the domain as follows:
\begin{align}
    \mathcal{D}_f =\{(\mathbf{x}_f^i,f(\mathbf{x}_f^i))\}_{i=1}^{N_f}= \{(\mathbf{x}_f^i,f^i)\}_{i=1}^{N_f}\\
    \mathcal{D}_b =\{(\mathbf{x}_b^i,b(\mathbf{x}_b^i))\}_{i=1}^{N_b} = \{(\mathbf{x}_b^i,b^i)\}_{i=1}^{N_b},
\end{align} 
with residual locations $\mathbf{x}_f^i\in \Omega$ and boundary locations $\mathbf{x}_b^i\in \partial\Omega$. With these notions, 
the corresponding PINN loss function is
defined as follows:

\begin{align}
    \mathcal{L}(\boldsymbol{\xi}) = \omega_u \mathcal{L}_u(\boldsymbol{\xi}) + \omega_f\mathcal{L}_f(\boldsymbol{\xi}) + \omega_b\mathcal{L}_b(\boldsymbol{\xi}),\label{al:pinnloss}
\end{align}
where 
\begin{align}
\mathcal{L}_u(\boldsymbol{\xi}) &= \frac{1}{N_u}\sum_{i=1}^{N_u}|u^i - \tilde{u}(\mathbf{x}_u^i;\boldsymbol{\theta})|^2,\\
\mathcal{L}_f(\boldsymbol{\xi}) &= \frac{1}{N_f}\sum_{i=1}^{N_f}|f^i - \mathcal{N}_x(\tilde{u}(\mathbf{x}_f^i;\boldsymbol{\theta});\boldsymbol{\lambda})|^2,\\
\mathcal{L}_b(\boldsymbol{\xi}) &= \frac{1}{N_b}\sum_{i=1}^{N_b}|b^i - \mathcal{B}_x(\tilde{u}(\mathbf{x}_b^i;\boldsymbol{\theta});\boldsymbol{\lambda})|^2,
\end{align}
and $\omega_u$, $\omega_f$, and $\omega_b$ are the weights for each term. In practice, this loss is often minimized with ADAM or L-BFGS optimizers. Standard PINNs typically provide only a deterministic estimate of the target parameters $\boldsymbol{\xi}$ \cite{RAISSI2019,Lu_2021,karniadakis2021physics}. These estimates may be inaccurate and unreliable for problems with small or noisy datasets. Therefore, qualifying the uncertainty in the estimate would be desirable. 

To account for the uncertainty, Bayesian Physics-Informed Neural Networks (B-PINNs) (e.g., \cite{sun2020, Yang_2021,LIN2022,antil2021novel}) have been proposed. B-PINNs are built on Bayesian Neural Networks (BNNs) by treating neural network weights and biases $\boldsymbol{\theta}$ and physical parameters $\boldsymbol{\lambda}$ as random variables. By  Bayes' theorem, the posterior distribution of the parameters $\mathbf{\xi}$ conditions on the forward measurements $\mathcal{D}_u$, the residual points $\mathcal{D}_f$, and the boundary points $\mathcal{D}_b$ can be obtained as follows:
\begin{align}
p(\boldsymbol{\xi}|\mathcal{D}_u,\mathcal{D}_f,\mathcal{D}_b) \propto p(\boldsymbol{\xi})p(\mathcal{D}_u,\mathcal{D}_f,\mathcal{D}_b|\boldsymbol{\xi}).
\end{align}
The choice of prior distribution $p(\boldsymbol{\xi})$ and likelihood $p(\mathcal{D}_u,\mathcal{D}_f,\mathcal{D}_b|\boldsymbol{\xi})$ will greatly affect the properties of the posterior distribution $p(\boldsymbol{\xi}|\mathcal{D}_u,\mathcal{D}_f,\mathcal{D}_b)$. A typical choice for the prior is to assume independence between the physical parameters $\boldsymbol{\lambda}$ and neural network parameters $\boldsymbol{\theta}$, i.e. $p(\boldsymbol{\xi}) = p(\boldsymbol{\theta})p(\boldsymbol{\lambda})$.
Additionally, the neural network parameters $\boldsymbol{\theta}=\{\theta^i\}_{i=1}^{N_\theta}$ are often assumed to follow independent zero-mean Gaussian distributions, i.e.
\begin{align}
p(\boldsymbol{\theta}) &= \prod_{i=1}^{N_\theta} p(\theta^i), \quad
p(\theta^i) \sim \mathcal{N}\left(0,\sigma^i_\theta\right),
\end{align}
where $\sigma^i_\theta$ is the standard deviation of the corresponding neural network parameter $\theta^i$.
For the likelihood, independence between the forward measurements $\mathcal{D}_u$, residual points $\mathcal{D}_f$, and boundary points $\mathcal{D}_b$ is often assumed as follows:
\begin{align}
p(\mathcal{D}_u,\mathcal{D}_f,\mathcal{D}_b|\boldsymbol{\xi}) &= p(\mathcal{D}_u|\boldsymbol{\xi})p(\mathcal{D}_f|\boldsymbol{\xi})p(\mathcal{D}_b|\boldsymbol{\xi}).
\end{align}
Each term within $\mathcal{D}_u$, $\mathcal{D}_f$, and $\mathcal{D}_b$ is often assumed to follow a Gaussian distribution of the form 
\begin{align}
\
p(\mathcal{D}_u|\boldsymbol{\xi}) &= \prod_{i=1}^{N_u}p(u^i|\boldsymbol{\xi}),\quad
p(\mathcal{D}_f|\boldsymbol{\xi}) = \prod_{i=1}^{N_f}p(f^i|\boldsymbol{\xi}),\quad
p(\mathcal{D}_b|\boldsymbol{\xi}) = \prod_{i=1}^{N_b}p(b^i|\boldsymbol{\xi}),\\
p(u^i|\boldsymbol{\xi})&=\frac{1}{\sqrt{2\pi\sigma_{\eta_u}^2}}\exp\left(-\frac{\left(u^i-\tilde{u}(\mathbf{x}_u^i;\boldsymbol{\theta})\right)^2}{2\sigma_{\eta_u}^2}\right),\\
p(f^i|\boldsymbol{\xi})&=\frac{1}{\sqrt{2\pi\sigma_{\eta_f}^2}}\exp\left(-\frac{\left(f^i-\mathcal{N}_x(\tilde{u}(\mathbf{x}_f^i;\boldsymbol{\theta});\boldsymbol{\lambda})\right)^2}{2\sigma_{\eta_f}^2}\right),\\
p(b^i|\boldsymbol{\xi})&=\frac{1}{\sqrt{2\pi\sigma_{\eta_b}^2}}\exp\left(-\frac{\left(b^i-\mathcal{B}_x(\tilde{u}(\mathbf{x}_b^i;\boldsymbol{\theta});\boldsymbol{\lambda})\right)^2}{2\sigma_{\eta_b}^2}\right).
\end{align}
Here, $\sigma_{\eta_u}$, $\sigma_{\eta_f}$, and $\sigma_{\eta_b}$ are the standard deviations for the forward measurement, residual point, and boundary point, respectively. Choice of physical parameter prior $p(\boldsymbol{\lambda})$ is often problem-dependent, as this distribution represents domain knowledge of the corresponding physical property. Given these choices of prior and likelihood functions, one can construct the corresponding posterior distribution of the BNN. For most BNNs, the closed-form expressions for the posterior distribution are unavailable, and approximate inference methods must be employed. Due to the overparameterized nature of neural networks, the resulting Bayesian inference problem is often high-dimensional for even moderately sized BNNs.

\subsection{Hamiltonian Monte Carlo (HMC)}
\label{subsec:HMC}
Next, we briefly review Hamiltonian Monte Carlo (HMC), a popular inference algorithm for B-PINNs \cite{Yang_2021} that serves as a baseline method for our proposed method. Hamiltonian Monte Carlo (HMC) is a powerful method for sampling based posterior inference \cite{DUANE1987216} and has been utilized in the context of inference in Bayesian Neural Networks (BNNs) \cite{neal2012bayesian}. HMC employs Hamiltonian dynamics to propose states in parameter space with high acceptance in the Metropolis-Hastings acceptance step. Given the posterior distribution $p(\boldsymbol{\xi}|\mathcal{D}_u,\mathcal{D}_f,\mathcal{D}_b)=e^{-U(\boldsymbol{\xi})}$, where $U$ is the negative log-density of the posterior, we define the Hamiltonian dynamics as follows
\begin{align}
H(\boldsymbol{\xi},\boldsymbol{r}) = U(\boldsymbol{\xi}) + \frac{1}{2}\boldsymbol{r}^TM^{-1}\boldsymbol{r},
\end{align}
where $\boldsymbol{r}\in\mathbb{R}^{N_\xi}$ is an auxiliary momentum vector, and $M\in\mathbb{R}^{N_\xi\times N_\xi}$ is the corresponding mass matrix, often set to the identity $I_{N_\xi}$. Starting from an initial sample of $\boldsymbol{\xi}$, the HMC generates proposal samples by resampling momentum $\boldsymbol{r}\sim\mathcal{N}(0,M)$ and advancing $(\boldsymbol{\xi},\boldsymbol{r})$ through Hamiltonian dynamics
\begin{align}
\frac{d\boldsymbol{\xi}}{dt} &= -M\boldsymbol{r},\\
\frac{d\boldsymbol{r}}{dt} &= -\nabla U(\boldsymbol{\xi}).
\end{align}
This is often done via Leapfrog integration \cite{DUANE1987216} for $L$ steps given a step size $\delta t$. Following this, a Metropolis-Hastings acceptance step is applied to determine if the given sample will be accepted. The details of the HMC are shown in Algorithm \ref{alg:HMC}. The variant HMC B-PINN used in this paper is based on the version in \cite{zou2022neuraluq}, to which we refer interested readers for more details.

\begin{algorithm}
\caption{Hamiltonian Monte Carlo (HMC)}\label{alg:HMC}
\begin{algorithmic}[1]
\State Input: $\boldsymbol{\xi}_0$ (initial sample), $\delta t$ (step size), $L$ (leapfrog steps)
\For{$i = 1,...,J$}
\State  $\boldsymbol{\xi}_{i} \leftarrow \boldsymbol{\xi}_{i-1}$
\State  Sample $\boldsymbol{r}_{i} \sim\mathcal{N}(0,M)$
\State  $\hat{\boldsymbol{\xi}}_{i} \leftarrow \boldsymbol{\xi}_{i}$
\State  $\hat{\boldsymbol{r}}_{i} \leftarrow \boldsymbol{r}_{i}$
\For{$j = 1,...,L$}
\State  $\hat{\boldsymbol{r}}_{i} \leftarrow \hat{\boldsymbol{r}}_{i} - \frac{\delta t}{2}\nabla U(\hat{\boldsymbol{\xi}}_{i})$
\State  $\hat{\boldsymbol{\xi}}_{i} \leftarrow \hat{\boldsymbol{\xi}}_{i} + \delta tM^{-1}\hat{\boldsymbol{r}}_{i}$
\State  $\hat{\boldsymbol{r}}_{i} \leftarrow \hat{\boldsymbol{r}}_{i} - \frac{\delta t}{2}\nabla U(\hat{\boldsymbol{\xi}}_{i})$
\EndFor
\State Sample $p\sim\mathcal{U}(0,1)$
\State $\alpha \leftarrow\min[1,\exp(H(\hat{\boldsymbol{\xi}}_{i},\hat{\boldsymbol{r}}_{i}) - H(\boldsymbol{\xi}_{i},\boldsymbol{r}_{i}))]$
\If{$p < \alpha$}
\State $\boldsymbol{\xi}_{i} \leftarrow \hat{\boldsymbol{\xi}}_{i}$
\EndIf
\EndFor
\State Return: $\boldsymbol{\xi}_{1},...,\boldsymbol{\xi}_{J}$ (Posterior samples)
\end{algorithmic}
\end{algorithm}

From the HMC algorithm, we obtain a set of approximate samples from the B-PINNs posterior distribution $p(\boldsymbol{\xi}|\mathcal{D}_u,\mathcal{D}_f,\mathcal{D}_b)$. We shall use these samples to obtain uncertainty estimates of the approximate forward solution and physical parameters. Denote $\bar{\lambda}$ and $\bar{{u}}$ to be the sample mean of physical parameter $\lambda$ and forward surrogate $\tilde{u}$, respectively. Additionally, we denote the corresponding sample standard deviations $s_{\lambda}$ and $s_{\tilde{u}}$. We compute the sample statistics over the $J$ samples $\{\lambda_{j}\}_{j=1}^J$ and $\{\tilde{u}(\mathbf{x};\boldsymbol{\theta}_{j})\}_{j=1}^J$ obtain from the HMC as follows:
\begin{align}
\bar{\lambda} &= \frac{1}{J}\sum_{j=1}^J \lambda_{j}, \quad
\bar{{u}}(\mathbf{x}) = \frac{1}{J}\sum_{j=1}^J\tilde{u}(\mathbf{x};\boldsymbol{\theta}_{j}),\label{al:m1}\\
s_{\lambda} &= \sqrt{\frac{\sum_{j=1}^J \left(\lambda_{j}-\bar{\lambda} \right)^2}{J-1}}, \quad
s_{\tilde{u}}(\mathbf{x}) = \sqrt{\frac{\sum_{j=1}^J \left(\tilde{u}(\mathbf{x};\boldsymbol{\theta}_{j})-\bar{u}(\mathbf{x}) \right)^2}{J-1}}\label{al:s2}.
\end{align}

\section{Ensemble Kalman Inversion-based B-PINNs}
\label{sec:EKI-section}
In this section, we first briefly review Ensemble Kalman Inversion (EKI) as an efficient method for solving Bayesian inverse problems. Following that, we present our proposed method, denoted {\it EKI B-PINNs}. Specifically, we first recast the setup of B-PINNs in the traditional Bayesian inverse problem setting and then employ EKI for efficient sampling-based inference.

\subsection{Ensemble Kalman Inversion (EKI)}
\label{subsec:EKI}
Ensemble Kalman Inversion (EKI) \cite{Iglesias_2013, iglesias_2015, Iglesias_2016, Kovachki_2019} is a popular class of methods that utilize the Ensemble Kalman Filter (EnKF) \cite{evensen2003ensemble} in the context of traditional inverse problems. These methods are derivative-free, easily parallelizable, and scale well in high-dimension inverse problems with ensemble sizes much smaller than the total number of parameters \cite{Kovachki_2019}. Assume that the unknown parameters $\boldsymbol{\xi}\in\mathbb{R}^{N_\xi}$ have a prior distribution $p(\boldsymbol{\xi})$ and the observations $\mathbf{y}\in\mathbb{R}^{N_y}$ are related to the parameters through the observation operator $\mathcal{G}$:
\begin{align}
\mathbf{y} = \mathcal{G}(\boldsymbol{\xi}) + \boldsymbol{\eta}\label{al:bip_1},
\end{align}
where $\boldsymbol{\eta}\in\mathbb{R}^{N_y}$ is a zero-mean Gaussian random vector with observation covariance matrix $R\in\mathbb{R}^{N_y\times N_y}$, i.e., $\boldsymbol{\eta}\sim\mathcal{N}(0,R)$. In the Bayesian context, this problem corresponds to the following posterior distribution: 
\begin{align}
p(\boldsymbol{\xi}|\mathbf{y}) &\propto p(\boldsymbol{\xi})\exp\left(-\frac{\left\|R^{-1/2}(\mathbf{y}-\mathcal{G}(\boldsymbol{\xi}))\right\|_2^2}{2}\right)\label{al:bip}.
\end{align}
 Given a prior of the form $p(\boldsymbol{\xi})\sim\mathcal{N}(\boldsymbol{\xi}_0,C_0)$, the posterior becomes
\begin{align}
p(\boldsymbol{\xi}|\mathbf{y}) \propto \exp\left(-\frac{\left\|C_0^{-1/2}(\boldsymbol{\xi}_0-\boldsymbol{\xi})\right\|_2^2 + \Big\|R_{}^{-1/2}(\mathbf{y}-\mathcal{G}(\boldsymbol{\xi}))\Big\|_2^2}{2}\right)
\label{al:bip_2},
\end{align}
For weakly nonlinear systems, approximate samples from \eqref{al:bip_2} can be obtained by minimizing an ensemble of loss functions of the form
\begin{align}
f(\boldsymbol{\xi}|\boldsymbol{\xi}_j,\mathbf{y}_j) = \frac{1}{2}\left\|C_0^{-1/2}(\boldsymbol{\xi}_j-\boldsymbol{\xi})\right\|_2^2 + \frac{1}{2}\Big\|R^{-1/2}(\mathbf{y}_j-\mathcal{G}(\boldsymbol{\xi}))\Big\|_2^2
\label{al:bip_3},
\end{align}
where $\boldsymbol{\xi}_j\sim \mathcal{N}(\boldsymbol{\xi}_0,C_0)$ and $\mathbf{y}_j\sim \mathcal{N}(\mathbf{y},R)$ \cite{chen2012ensemble,evensen2022}. EnKF-based methods such as EKI can be derived as an approximation to the minimizer of an ensemble of loss functions of the form \eqref{al:bip_3} \cite{evensen2022}. 

In practice, EKI and its variants consider the following artificial dynamics state-space model formulation based on the original Bayesian inverse problem \eqref{al:bip_1} so that the EnKF update equations can be applied:
\begin{align}
\boldsymbol{\xi}_i&=\boldsymbol{\xi}_{i-1} + \boldsymbol{\epsilon}_i,\quad \boldsymbol{\epsilon}_i\sim \mathcal{N}(0,Q),\label{al:ad1}\\
\mathbf{y}_i&=\mathcal{G}(\boldsymbol{\xi}_{i}) + \boldsymbol{\eta}_i,\quad \boldsymbol{\eta}_i\sim \mathcal{N}(0,R),\label{al:ad2}
\end{align}
where $\boldsymbol{\epsilon}_i$ is an artificial parameter noise term with the artificial parameter covariance $Q\in\mathbb{R}^{N_\xi\times N_\xi}$ and $\boldsymbol{\eta}_i$ represents the observation error with the observation covariance $R\in\mathbb{R}^{N_y\times N_y}$. Given an initial ensemble of $J$ ensemble members $\{\boldsymbol{\xi}_0^{(j)}\}_{j=1}^J$, the iterative EKI methods iteratively correct the ensemble $\{\boldsymbol{\xi}_i^{(j)}\}_{j=1}^J$ based on Kalman update equations similar to \cite{Kovachki_2019}:
\begin{align}
\hat{\boldsymbol{\xi}}_i^{(j)} &= \boldsymbol{\xi}_{i-1}^{(j)} + \boldsymbol{\epsilon}_i^{(j)}, \quad \boldsymbol{\epsilon}_i^{(j)}\sim \mathcal{N}(0,Q),\label{al:EKI-2-1}\\
\hat{\mathbf{y}}^{(j)}_i&=\mathcal{G}(\hat{\boldsymbol{\xi}}_i^{(j)}),\label{al:EKI-2-2}\\
\boldsymbol{\xi}_i^{(j)} &= \hat{\boldsymbol{\xi}}_i^{(j)} + {C}_{i}^{\hat{\xi} \hat{y}}({C}^{\hat{y}\hat{y}}_i + R)^{-1}(\mathbf{y}-\hat{\mathbf{y}}^{(j)}_{i}+\boldsymbol{\eta}_i^{(j)}), \quad \boldsymbol{\eta}_i^{(j)}\sim \mathcal{N}(0,R),\label{al:EKI-2-3}
\end{align}
where ${C}^{\hat{y}\hat{y}}_i$ and ${C}^{\hat{\xi}\hat{y}}_i$ are the sample covariance matrices defined as follows: 
\begin{align}
{C}^{\hat{y}\hat{y}}_i &= \frac{1}{J-1}\sum_{j=1}^J (\hat{\mathbf{y}}^{(j)}_{i}-\bar{\mathbf{y}}_i)(\hat{\mathbf{y}}^{(j)}_{i}-\bar{\mathbf{y}}_i)^T\label{al:EKI-2-4},\\
{C}^{\hat{\xi}\hat{y}}_i &= \frac{1}{J-1}\sum_{j=1}^J (\hat{\boldsymbol{\xi}}_{i}^{(j)}-\bar{\boldsymbol{\xi}}_i)(\hat{\mathbf{y}}_{i}^{(j)}-\bar{\mathbf{y}}_i)^T.\label{al:EKI-2-5}
\end{align} 
Here, $\bar{\boldsymbol{\xi}}_i$ and $\bar{\mathbf{y}}_i$ are the corresponding sample average of prior ensembles $\{\hat{\boldsymbol{\xi}}^{(j)}_i\}$ and $\{\hat{{\mathbf{y}}}^{(j)}_i\}$.
We remark that for many EKI variants, $Q=0$, which may lead to the ensemble collapsing \cite{Chada_2019}. This ensemble collapse is not observed for positive definite observation error covariance $Q$, and desirable convergence properties with reasonable uncertainty estimates have been shown for traditional Bayesian inverse problems \cite{huang2022iterated, huang2022}. In the case of Gaussian prior and linear measurement operator, convergence to the correct Bayesian posterior corresponding to the artificial dynamics \eqref{al:ad1}-\eqref{al:ad2} can be shown \cite{Gland2009}. However, these assumptions are typically not satisfied, and thus the corresponding estimates will be biased. Nonetheless, empirical evidence suggested that EKI can still provide reasonable posterior estimates even when these assumptions are violated \cite{Botha2022}.

\begin{algorithm}
\caption{Ensemble Kalman Inversion (EKI)}\label{alg:eki}
\begin{algorithmic}[1]
\State Input: $\mathbf{y}$ (observations), $Q$ (evolution covariance), $R$ (observation covariance)
\State Initialize prior samples for $j=1,...,J$:
\begin{align*}
\boldsymbol{\xi}_0^{(j)}&\sim p(\boldsymbol{\xi}).
\end{align*}
\For{$i=1,...,I$}
\State Obtain $i$-th prior parameter and measurement ensembles for $j=1,...,J$:
\begin{align*}
\boldsymbol{\epsilon}_i^{(j)}&\sim \mathcal{N}(0,Q), \quad
\boldsymbol{\eta}_i^{(j)}\sim \mathcal{N}(0,R).\\
\hat{\boldsymbol{\xi}}_i^{(j)}&=\boldsymbol{\xi}_i^{(j)}+\boldsymbol{\epsilon}_i^{(j)}.\\
\hat{\mathbf{y}}_i^{(j)}&=\mathcal{G}_i(\hat{\boldsymbol{\xi}}_i^{(j)}).
\end{align*}
\State Compute the sample mean and covariance:
\begin{align*}
\bar{\boldsymbol{\xi}}_i &= \frac{1}{J}\sum_{j=1}^J \hat{\boldsymbol{\xi}}_{i}^{(j)}, \quad
\bar{\mathbf{y}}_i = \frac{1}{J}\sum_{j=1}^J \hat{\mathbf{y}}_{i}^{(j)}.\\
{C}^{\hat{y}\hat{y}}_i &= \frac{1}{J-1}\sum_{j=1}^J (\hat{\mathbf{y}}^{(j)}_{i}-\bar{\mathbf{y}}_i)(\hat{\mathbf{y}}^{(j)}_{i}-\bar{\mathbf{y}}_i)^T.\\
{C}^{\hat{\xi}\hat{y}}_i &= \frac{1}{J-1}\sum_{j=1}^J (\hat{\boldsymbol{\xi}}_{i}^{(j)}-\bar{\boldsymbol{\xi}}_i)(\hat{\mathbf{y}}_{i}^{(j)}-\bar{\mathbf{y}}_i)^T.
\end{align*}
\State Update the posterior ensemble for $j=1,...,J$: 
\begin{align*}
\boldsymbol{\xi}_i^{(j)} &= \hat{\boldsymbol{\xi}}_i^{(j)} + {C}^{\hat{\xi}\hat{y}}_i({C}^{\hat{y}\hat{y}}_i + R)^{-1}(\mathbf{y}-\hat{\mathbf{y}}_i^{(j)} + \boldsymbol{\eta}_i^{(j)}).
\end{align*}
\EndFor
\State Return: $\boldsymbol{\xi}_I^{(1)},...,\boldsymbol{\xi}_I^{(J)}$
\end{algorithmic}
\end{algorithm}

\subsection{EKI B-PINN}
\label{sec:EKI-PINN}
While EKI-based methods have often been applied to estimate physical model parameters (e.g., \cite{Iglesias_2013, chada2018parameterizations, ChadaTik_2019}) under the context of traditional Bayesian inverse problems, these approaches often rely on existing numerical forward solvers or the corresponding discrete operators. As the EKI requires multiple evaluations of the numerical forward solvers over EKI iterations, this can be computationally expensive for large-scale complex applications. In contrast, our approach (EKI B-PINN) learns a neural network surrogate and infers the model parameters simultaneously without the need for a traditional numerical solver. Combining the inexpensive forward surrogate with EKI allows us to efficiently explore the high-dimensional posterior with larger ensemble sizes.

To employ EKI under the context of B-PINNs, we first recast the setup of B-PINNs in section \ref{subsec:B-PINN} by interpreting the corresponding notation in the context of EKI. Recall the notation introduced in B-PINNs section \ref{subsec:B-PINN}: the forward measurement vector $\mathbf{u}=\{u^i\}_{i=1}^{N_u}$, the residual vector $\mathbf{f}=\{f^i\}_{i=1}^{N_f}$ and boundary vector $\mathbf{b}=\{b^i\}_{i=1}^{N_b}$ from datasets $\mathcal{D}_u=\{(\mathbf{x}_u^i,u^i)\}_{i=1}^{N_u}$, $\mathcal{D}_f=\{(\mathbf{x}_f^i,f^i)\}_{i=1}^{N_f}$  and $\mathcal{D}_b=\{(\mathbf{x}_b^i,b^i)\}_{i=1}^{N_b}$. We utilize the concatenated physical and neural network parameters $\boldsymbol{\xi}=[\boldsymbol{\lambda},\boldsymbol{\theta}]$. With the approximate solution $\tilde{u}(\mathbf{x};\boldsymbol{\theta})$ parameterized by $\boldsymbol{\theta}$, the PDE operator $\mathcal{N}_x$, and the boundary operator $\mathcal{B}_x$, we can define the corresponding observation operator $\mathcal{G}_u$, $\mathcal{G}_f$, and $\mathcal{G}_b$:
\begin{align}
\mathcal{G}_u(\boldsymbol{\xi}) &= [\tilde{u}(\mathbf{x}_u^1;\boldsymbol{\theta}),...,\tilde{u}(\mathbf{x}_u^{N_u};\boldsymbol{\theta})],\\
\mathcal{G}_f(\boldsymbol{\xi}) &= [\mathcal{N}_x(\tilde{u}(\mathbf{x}_f^1;\boldsymbol{\theta});\boldsymbol{\lambda}),...,\mathcal{N}_x(\tilde{u}(\mathbf{x}_f^{N_f};\boldsymbol{\theta});\boldsymbol{\lambda})],\\
\mathcal{G}_b(\boldsymbol{\xi}) &= [\mathcal{B}_x(\tilde{u}(\mathbf{x}_u^1;\boldsymbol{\theta});\boldsymbol{\lambda}),...,\mathcal{B}_x(\tilde{u}(\mathbf{x}_b^{N_b};\boldsymbol{\theta});\boldsymbol{\lambda})].
\end{align}
Given these notations, we now define our measurement vector $\mathbf{y}$ and the corresponding measurement operator $G(\boldsymbol{\xi})$ under the context of EKI as follows:
\begin{align}
\mathbf{y} &=[\mathbf{u},\mathbf{f},\mathbf{b}],\\
\mathcal{G}(\boldsymbol{\xi}) &= [\mathcal{G}_u(\boldsymbol{\xi}), \mathcal{G}_f(\boldsymbol{\xi}), \mathcal{G}_b(\boldsymbol{\xi})].
\end{align}
After identifying each component in the EKI setting, we employ the version of EKI in \eqref{al:EKI-2-1}-\eqref{al:EKI-2-3} to infer the parameters of B-PINNs. Following that, we can then compute the sample means and standard deviations of ensemble $\{\lambda^{(j)}\}_{i=1}^J$ of the physical parameter $\mathcal{\lambda}$ and ensemble of approximate solutions $\{\tilde{u}(\mathbf{x};\boldsymbol{\theta^{(j)}})\}_{j=1}^J$ via \eqref{al:m1}-\eqref{al:s2} as described in the section \ref{subsec:HMC}.

We remark that in EKI B-PINN, the underlying physics laws are enforced as soft constraints on the residual points. This approach contrasts with the standard EKI measurement operator, which enforces the physics law as hard constraints imposed by the traditional numerical solvers. 

\subsection*{Choice of Covariance matrices $Q$ and $R$}
The choice of observation covariance $R$ and parameter evolution $Q$ are often important in the Ensemble Kalman type algorithms, including the EKI algorithms. In the Ensemble Kalman methods literature, several attempts to automatically estimate these covariance matrices have also been suggested (e.g., \cite{ Berry2013, li2009simultaneous, Anderson2007}), however, there is no standard approach to estimating these matrices. In practice, $R$ is often assumed to be known or estimated empirically from the instrument error and representation error between the states and observations \cite{tandeo2018}. The matrix $Q$ is more difficult to estimate, as its dimension is often much larger than the number of available measurements.

In this paper, we assume $R$ corresponds to the covariance matrix chosen for the Gaussian likelihood $p(\mathbf{y}|\boldsymbol{\xi})$ in Section \ref{subsec:B-PINN}, i.e.,
\begin{align}
R &= \begin{bmatrix}\sigma_{\eta_u}^2I_{N_u}&0&0\\0&\sigma_{\eta_f}^2I_{N_f}&0\\0&0& \sigma_{\eta_b}^2I_{N_b} \end{bmatrix}.
\label{al:R}
\end{align}
The choice of $Q$ is important in this setting as it prevents the collapse of the ensemble and improves the uncertainty quantification in the EKI. In this study, we assume $Q$ takes the form
\begin{align}
Q = \begin{bmatrix}
    \sigma^2_\theta I_{N_\theta}&0\\
    0&\sigma^2_\lambda I_{N_\lambda}
\end{bmatrix}, \label{al:Q}
\end{align}
where $\sigma_\theta$ and $\sigma_\lambda$ are standard deviations associated with the neural weight parameters $\boldsymbol{\theta}$ and the physical parameters $\boldsymbol{\lambda}$, respectively. 

\subsection*{Stopping Criterion}
We consider a stopping criterion based on the Discrepancy Principle \cite{iglesias_2015, Iglesias_2013, Hanke1997, morozov2012methods}, originally utilized as a stopping criterion in iterative regularized Gauss-Newton methods, which has been used as a common choice in many EKI formulations. The discrepancy principle suggests an acceptable choice of $\boldsymbol{\xi}$ for the inverse problem \eqref{al:bip_1} can be made when
\begin{align}
||R^{-1/2}(y-\mathcal{G}(\boldsymbol{\xi}))|| \leq ||R^{-1/2}(y-\mathcal{G}(\boldsymbol{\xi}^\dagger))||,
\end{align}
where $\boldsymbol{\xi}^{\dagger}$ is the true solution of \eqref{al:bip_1}. This choice avoids instabilities in the solution and provides a criterion for stopping the EKI iteration. As the $\boldsymbol{\xi}^{\dagger}$ is not known, often it is stated as 
\begin{align}
||R^{-1/2}(y-G(\boldsymbol{\xi}))|| \leq \eta
\end{align}
where $\eta>0$ is some stopping threshold. In the context of EKI, a sample mean-based discrepancy principle stopping criterion has been employed \cite{iglesias_2015}:
\begin{align}
\left\Vert R^{-1/2}\left(y-\frac{1}{J}\sum_{j=1}^{J}\mathcal{G}(\boldsymbol{\xi^{(j)}_i})\right)\right\Vert \leq \eta.
\end{align}
The choice of $\eta$ is dependent on $R$, which can be an issue if a reasonable choice of $R$ is not clear, such as in the case with residual points $\mathcal{D}_f$ and boundary points $\mathcal{D}_b$ in the context of B-PINNs. Instead, we consider the relative change in the discrepancy over several iterations. By defining $D_i$ as the $i$th discrepancy metric as follows:
\begin{align}
D_i = \left\Vert R^{-1/2}\left(y-\frac{1}{J}\sum_{j=1}^{J}\mathcal{G}(\boldsymbol{\xi^{(j)}_i})\right)\right\Vert \label{al:sc},
\end{align}
We shall stop at the iteration when the relative improvement of $D_i$ over a fixed iteration window of length $W$ does not improve by more than $\tau$, i.e.,
\begin{align}
\underset{j\in\{i-W,...,i\}}{\max}\frac{|D_j-D_i|}{D_i}<\tau.\label{al:sc_p}
\end{align}

\begin{remark}
If an EKI iteration results in a failure state which occurs in the first several iterations, we can reinitialize the ensemble from the prior $\boldsymbol{\xi}_i^{(j)}\sim p(\boldsymbol{\xi})$.
\end{remark}
\subsection*{Complexity Analysis}

At each EKI iteration in Algorithm \ref{alg:eki} has following computational complexity of $\mathcal{O}(JN_y N_\xi + N_y^3+JN_y^2)$ detailed as follows:
\begin{itemize}
\item $\mathcal{O}(JN_y^2)$ -  Construction of matrix $C^{\hat{y}\hat{y}}$ in \eqref{al:EKI-2-4}
\item $\mathcal{O}(JN_{\xi}N_y)$ -  Construction of matrix $C^{\hat{\xi}\hat{y}}$ in \eqref{al:EKI-2-5}
\item $\mathcal{O}(N_y^3 + JN_y^2 + JN_{\xi}N_y)$ - Evaluation of $C^{\hat{\xi}\hat{y}}(C^{\hat{y}\hat{y}}+ R)^{-1}(\hat{\mathbf{y}}_i^{(j)}+\boldsymbol{\eta}_i^{(j)})$ in \eqref{al:EKI-2-3}.
\end{itemize}
As a result, the computational complexity of EKI grows linearly with both parameter dimension and ensemble size and thus can easily scale to high-dimensional inverse problems efficiently. Nevertheless, the method does scale cubically with data dimension and hence is computationally prohibitive when naively applied to large data sets. In this scenario, mini-batching, as in standard neural network optimization problems, can be used to reduce the computational cost \cite{Kovachki_2019}. Furthermore, it is important to note that the EKI requires storing an ensemble of $J$ neural network parameter sets, which can be memory-demanding for large ensemble sizes and large networks. Dimension reduction techniques might be helpful to address this issue \cite{chen2023reduced}.

\begin{algorithm}
\caption{Ensemble Kalman Inversion (EKI)  B-PINNs}\label{alg:eki_2}
\begin{algorithmic}[1]
\State Input: $\mathbf{y}$ (observations), $Q$ (parameter covariance), $R$ (observation covariance),  $W$ (stopping window), $\tau$ (stopping threshold)
\State Initialize prior samples for $j=1,...,J$:
\begin{align*}
\boldsymbol{\xi}_0^{(j)}&\sim p(\boldsymbol{\xi}).
\end{align*}
\For{$i=1,...,I$}
\State Obtain $i$ th prior parameter and measurement ensembles for $j=1,...,J$:
\begin{align*}
\boldsymbol{\epsilon}_i^{(j)}&\sim \mathcal{N}(0,Q).\\
\hat{\boldsymbol{\xi}}_i^{(j)}&=\boldsymbol{\xi}_i^{(j)}+\boldsymbol{\epsilon}_i^{(j)}.\\
\hat{\mathbf{y}}_i^{(j)}&=\mathcal{G}_i(\hat{\boldsymbol{\xi}}_i^{(j)}).
\end{align*}
\State Evaluate sample mean and covariance terms:
\begin{align*}
\bar{\boldsymbol{\xi}}_i &= \frac{1}{J}\sum_{j=1}^J \hat{\boldsymbol{\xi}}_{i}^{(j)}.\\
\bar{\mathbf{y}}_i &= \frac{1}{J}\sum_{j=1}^J \hat{\mathbf{y}}_{i}^{(j)}.\\
{C}^{\hat{y}\hat{y}}_i &= \frac{1}{J-1}\sum_{j=1}^J (\hat{\mathbf{y}}^{(j)}_{i}-\bar{\mathbf{y}}_i)(\hat{\mathbf{y}}^{(j)}_{i}-\bar{\mathbf{y}}_i)^T.\\
{C}^{\hat{\xi}\hat{y}}_i &= \frac{1}{J-1}\sum_{j=1}^J (\hat{\boldsymbol{\xi}}_{i}^{(j)}-\bar{\boldsymbol{\xi}}_i)(\hat{\mathbf{y}}_{i}^{(j)}-\bar{\mathbf{y}}_i)^T.
\end{align*}
\State Update posterior ensemble for $j=1,...,J$: 
\begin{align*}
\boldsymbol{\eta}_i^{(j)}&\sim \mathcal{N}(0,R),\\
\boldsymbol{\xi}_i^{(j)} &= \hat{\boldsymbol{\xi}}_i^{(j)} + {C}_{i}^{\hat{\xi} y}({C}^{yy}_i + R)^{-1}(\mathbf{y}-\hat{\mathbf{y}}_i^{j}+\boldsymbol{\eta}_i^{(j)}).
\end{align*}
\State Check discrepancy
\begin{align*}
    D_i &= \left\Vert R^{1/2}\left(y-\frac{1}{J}\sum_{j=1}^{J}\mathcal{G}(\boldsymbol{\xi^{(j)}_i})\right)\right\Vert,
\end{align*}
\If{$\underset{j\in\{i-W,...,i\}}{\max}|D_j-D_i|/D_i<\tau$}
\State $I := i$
\State \textbf{Break}
\EndIf
\EndFor
\State Return: $\boldsymbol{\xi}^{(1)}_I,..., \boldsymbol{\xi}^{(J)}_I$
\end{algorithmic}
\end{algorithm}
\section{Numerical Examples}
\label{sec:NumExample}

In this section, we shall demonstrate the applicability and performance of EKI B-PINNs via various numerical examples. We also compare our approach with a variant of the HMC B-PINN to assess the inference accuracy and computational efficiency of our proposed method.

For each example, we generated a synthetic data set, $\mathcal{D}_u$, by solving the corresponding problem and corrupting the solution with i.i.d zero-mean Gaussian noise, $\mathcal{N}(0,\sigma_u)$. To demonstrate the robustness of the EKI method, we considered two noise levels: $\sigma_u=0.1$ and $\sigma_u=0.01$. We generate residual points $\mathcal{D}_f$ by evaluating $f$ at locations generated using Latin hypercube sampling over the problem domain for 2D problems and equally spaced over the domain for 1D problems. For 2D problems, boundary points $\mathcal{D}_b$ are placed equally spaced over the boundary. Boundary and residual points are assumed to be noise-free in this paper. 

For all examples, the neural network architecture used for both B-PINNs consists of 2 hidden layers with 50 neurons in each layer and the $\tanh$ activation function. Neural network parameter dimension $N_\theta$ for each type of example can be seen in Table~\ref{tab:param_num}. For each problem, we assume the standard deviations for the boundary and residual likelihood in \eqref{al:R}: $\sigma_{\eta_b}=0.01$ and $\sigma_{\eta_f}=0.01$ for the B-PINNs unless otherwise specified. Furthermore, we assume that the measurement noise level $\sigma_u$ is known for each example and set $\sigma_{\eta_u}=\sigma_u$. Finally, unless otherwise specified, we use physical parameter prior $\boldsymbol{\lambda}\sim\mathcal{N}(0,I_{N_\lambda})$ for both B-PINNs.

For the EKI B-PINNs, we choose an ensemble size of $J=1000$ and stopping criterion parameters from \eqref{al:sc_p}: $W=25$ and $\tau=0.05$. Additionally, the artificial dynamics standard deviations for the parameters from \eqref{al:Q} are chosen to be $\sigma_\lambda=0.1$ and $\sigma_\theta=0.002$. For the HMC B-PINNs, the leapfrog step $L=50$ and the initial step size $\delta t=0.1$ adaptively tuned to reach an acceptance rate of $60\%$ during burn-in steps as in \cite{zou2022neuraluq}. We draw a total of 1000 samples following 1000 burn-in steps.

\begin{table}[]
\centering
\begin{tabular}{|l|l|}
\hline
& $\boldsymbol{\theta}$ size \\ \hline
1D PDE - Examples \ref{subsec:Ex0}, \ref{subsec:Ex1} & 5251  \\ \hline
2D PDE - Examples \ref{subsec:Ex2}, \ref{subsec:Ex6} \ref{subsec:Ex8}  & 5301 \\ \hline
System of ODEs - Example \ref{subsec:Ex5} & 5353 \\ \hline
\end{tabular}
\caption{The number of neural network parameters $\boldsymbol{\theta}$ for each example. \label{tab:param_num}}
\end{table}

\subsection*{Metrics} 
We examine the accuracy of the forward solution approximation and physical parameter estimation from the B-PINNs in the following metrics over an independent test set $\{\mathbf{x}_t^i\}_{i=1}^{N_t}$:
\begin{equation}
\begin{split}
& e_{u} =\sqrt{\frac{\sum_{i=1}^{N_t}\left| u(\mathbf{x}_t^i) - \bar{u}(\mathbf{x}_t^i)\right|^2}{\sum_{i=1}^{N_t}\left|u(\mathbf{x}_t^i)\right|^2}},  
\quad e_{{\lambda}} = \frac{\left| \lambda - \bar{\lambda} \right|}{\left|\lambda\right|},
\end{split}
\end{equation}
where $u(x)$ and $\lambda$ are the reference solution and the reference physical parameters.
The sample means of the forward solution approximation $\bar{u}$ and the physical parameter $\bar{\lambda}$ are computed from the B-PINNs as defined in \eqref{al:m1}. The mean predictions are obtained in the final iteration for the EKI B-PINNs and over the 1000 samples generated for the HMC B-PINN. Furthermore, we assess the quality of our uncertainty estimates by examining the sample standard deviation of the estimated forward solutions and physical parameters $\sigma_\lambda$ and $\sigma_{\tilde{u}}$ defined in \eqref{al:s2}.  

To demonstrate the efficiency of our proposed method, we compare the walltime of the EKI B-PINN and HMC B-PINN (including both burn-in and training time) experiments averaged over 10 trials. The average iterations utilized over 10 trials are also presented for the EKI B-PINN tests. All experiments use the JAX \cite{jax2018github} library in single-precision floating-point on a single GPU (Tesla T4 GPU with 16 GB of memory).

\subsection{1D linear Poisson equation}
\label{subsec:Ex0}
We first consider the 1D linear Poisson equation motivated by \cite{ceccarelli2021bayesian} as follows:
\begin{align}
    u_{xx} - k\cos(x)=0,\quad x\in[0,8],\label{al:Ex0}\\
    u(0)=0,\quad u(8)=k\cos(8),
\end{align}
where the exact solution is $u(x)=k\cos(x)$. Assuming the unknown constant $k=1.0$, we generated $N_u=8$ equally spaced measurements over $[0,8]$ for measurements $\mathbf{u}$, excluding the boundary points, and $N_b=2$ boundary points at $x=0$ and $x=8$. Additionally, $N_f=100$ equally spaced residual points are utilized.

Due to the linearity of the solution with respect to $k$, one can derive a Gaussian ``reference" posterior distribution for parameter $k$ conditioned on knowledge of the correct solution parameterization $\tilde{u}(x;k)=k\cos(x)$ \cite{ceccarelli2021bayesian}. In the case where the forward solution and boundary standard deviations are equal (i.e., $\sigma_{\eta_u}=\sigma_{\eta_b}$), if we denote for simplicity $\mathcal{D}_u$ as containing both forward solution and boundary data, the distribution takes the form: 
\begin{align}
    p(k|\mathcal{D}_u) \propto \exp\left(-\left(\frac{\sum_{i=1}^{N_u} (u^i-k\cos(x_u^i))^2}{2\sigma_{\eta_{u}}^2} + \frac{(k-k_0)^2}{2\sigma_k^2}\right)\right),
\end{align}
where $k_0$, $\sigma_k$ are the prior mean and standard deviation of $k$.

\begin{figure}
\centering
\includegraphics[scale=1.0]{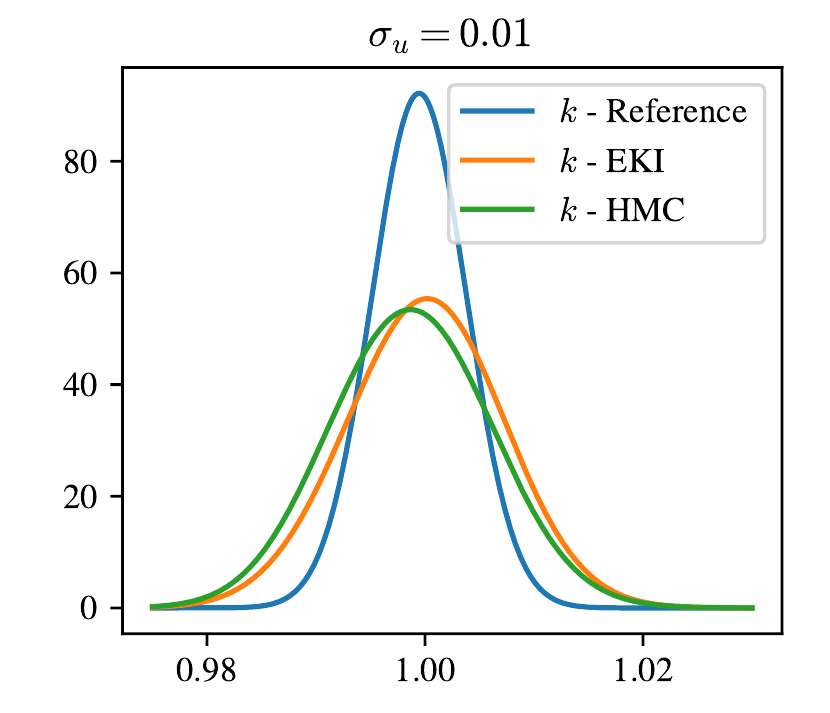}
\vspace{-0.0cm}
\caption{Example \ref{subsec:Ex0}: posterior distributions of $k$ for EKI B-PINN and HMC B-PINN at $\sigma_u=0.01$ with the corresponding reference posterior distribution. \label{fig:Ex_0_density}}
\end{figure}

In this example, we choose $k_0=0, \sigma_k=1$ and consider the noise level $\sigma_u=0.01$. We present the ``reference" density and density estimates of the estimated posterior of $k$ in Fig~\ref{fig:Ex_0_density}. Both approaches deliver comparable inference results for the posterior distribution of $k$ compared to the ``reference" density.

Table \ref{tab:ex0_confidence} provides the mean and one standard deviation of the posterior $k$ approximations. The uncertainty estimates of $k$ for both B-PINNs show that the true value of $k$ is within one standard deviation of the posterior mean. Table \ref{tab:ex0_error} shows the mean relative error of the solution and parameters, with corresponding walltime and the number of EKI iterations utilized. From the table, it is clear that the EKI B-PINN can achieve comparable approximation quality to HMC B-PINN but with a 8-fold speed-up.

\begin{table}
\centering
\begin{tabular}{c|c|c}
\hline
& &  $k \text{ (mean} \pm \text{std)}$  \\
\hline 
\multirow{ 2}{*}{$\sigma_u=0.01$}& EKI &  $1.002
 \pm 0.005$  \\
&HMC &  $1.001\pm 0.005$  \\
\hline
\end{tabular}
\caption{Example \ref{subsec:Ex0}: sample mean and standard deviation of parameter $k$ for EKI B-PINN and HMC B-PINN for $\sigma_u=0.01$ noise level. The true value of $k$ is 1.
\label{tab:ex0_confidence}} 
\end{table} 

\begin{table}
\centering
\begin{tabular}{c|c|cc|c|c}
\hline
& &  $e_{u}$  &  $e_{k}$  & Walltime & Iterations\\
\hline 
\multirow{ 2}{*}{$\sigma_u=0.01$}&EKI & 0.63\% & 0.18\%&5.81 seconds& 219\\
&HMC & 1.43\% & 0.08\%&  46.66 seconds&-\\
\hline
\end{tabular}
\caption{Example \ref{subsec:Ex0}: relative errors $e_u$ of the forward solution $u$ and $e_k$ of parameter $k$ for the noise level $\sigma_u=0.01$, as well as average walltime. Average EKI iterations for the EKI B-PINN are also reported.
\label{tab:ex0_error}} 
\end{table}

\subsection{1D nonlinear Poisson equation}
\label{subsec:Ex1}

\begin{table}
\centering
\begin{tabular}{c|c|c}
\hline
& &  $k \text{ (mean} \pm \text{std)}$  \\
\hline 
\multirow{ 2}{*}{$\sigma_u=0.01$}& EKI &  $0.701 \pm 0.006$  \\
&HMC &  $0.701 \pm 0.006$  \\
\hline
\multirow{ 2}{*}{$\sigma_u=0.1$}& EKI &  $0.697 \pm 0.012$  \\
&HMC &  $0.688 \pm 0.022$  \\ 
\hline
\end{tabular}
\caption{Example \ref{subsec:Ex1}: sample mean and standard deviation of parameter $k$ for EKI B-PINN and HMC B-PINN for $\sigma_u=0.01, 0.1$ noise levels. The true value of $k$ is 0.7.
\label{tab:ex1_confidence}} 
\end{table} 

\begin{figure}
\includegraphics[width=1\linewidth]{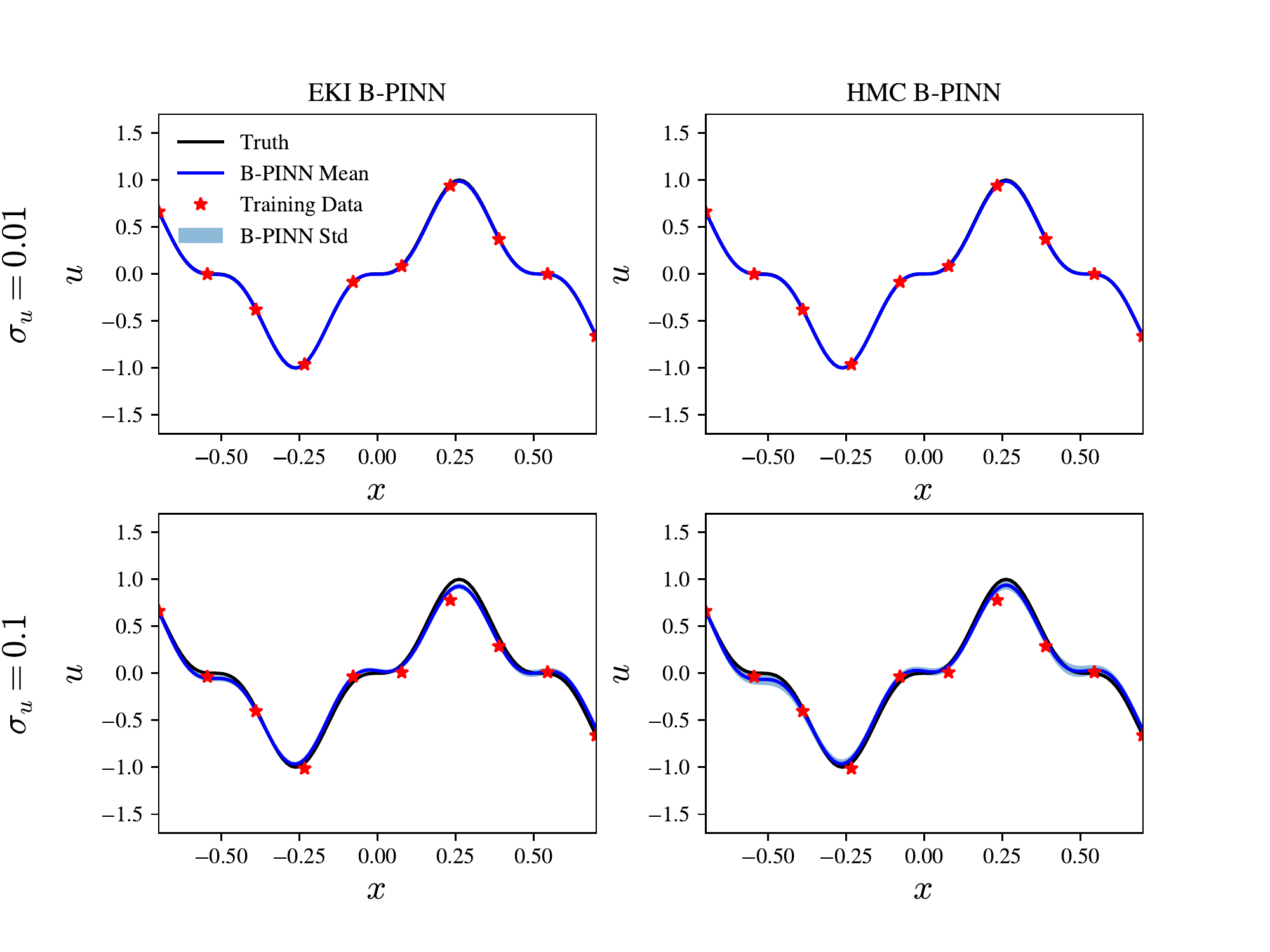}
\vspace{-1.0cm}
\caption{Example \ref{subsec:Ex1}: sample mean and standard deviation of $u$ for EKI B-PINN and HMC B-PINN for $\sigma_u=0.01, 0.1$ noise levels. \label{fig:Ex_1_u}}
\end{figure}

We now consider a 1D nonlinear Poisson equation as presented in \cite{Yang_2021} as follows:
\begin{align}
\lambda u_{xx} + k \tanh(u) &= f, \quad x \in [-0.7, 0.7], \label{al:ex1}
\end{align}
where $\lambda=0.01$. Here, $k = 0.7$ is the unknown physical parameter to be inferred. By assuming the true solution $u(x) = \sin^3(6x)$, the right-hand side $f$ and boundary conditions can be analytically constructed. The objective of this problem is to infer the physical parameter $k$ and the forward solution $u$, along with corresponding uncertainty estimates, using $N_u=6$ measurements $\mathbf{u}$ equally spaced over the spatial domain, $N_b=2$ boundary points $\mathbf{b}$, and $N_f=32$ equally spaced PDE residual points $\mathbf{f}$.

Table \ref{tab:ex1_confidence} compares the posterior sample mean and one standard deviation of $k$ obtained from the EKI B-PINN and HMC B-PINN. Both methods accurately capture the true value of $k = 0.7$ within one standard deviation of the mean for both noise levels. Figure \ref{fig:Ex_1_u} compares the sample mean and standard deviation of the surrogate based on both approaches. The mean of the EKI B-PINN is in good agreement with the reference solution. The one standard deviation bound captures most of the error in the mean prediction.

\begin{table}
\centering
\begin{tabular}{c|c|cc|c|c}
\hline
& &  $e_{u}$  &  $e_{k}$  & Walltime & Iterations\\
\hline 
\multirow{ 2}{*}{$\sigma_u=0.01$}&EKI &  1.19\% & 0.19\%&6.43 seconds& 282\\
&HMC & 1.12\% & 0.10\%& 55.55 seconds&-\\
\hline
\multirow{ 2}{*}{$\sigma_u=0.1$}&EKI & 9.32\% & 0.38\%&6.47 seconds&289\\
&HMC & 8.76\% & 1.73\%& 55.77 seconds&-\\ 
\hline
\end{tabular}
\caption{Example \ref{subsec:Ex1}: relative errors $e_u$ of the forward solution $u$ and $e_k$ of parameter $k$ for the noise levels $\sigma_u=0.01, 0.1$, as well as average walltime. Average EKI iterations for the EKI B-PINN are also reported.
\label{tab:ex1_error}} 
\end{table} 

Table \ref{tab:ex1_error} reports the relative errors of the forward solution and parameter estimates using EKI B-PINN and HMC B-PINN for two noise levels, along with the corresponding walltime and number of EKI iterations. Both methods are reasonably accurate, with estimates of $k$ within $1\%$ error for $\sigma_u = 0.01$ and $5\%$ error for $\sigma_u = 0.1$. Additionally, the relative errors of the mean forward solution $e_u$ for both B-PINNs reach similar levels of accuracy, approximately $2\%$ and $10\%$ for $\sigma_u=0.01$ and $\sigma_u=0.1$ respectively. Nevertheless, the EKI B-PINN achieves comparable inference results to the HMC B-PINN, but approximately 9 times faster.

\subsection{2D nonlinear diffusion-reaction equation}
\label{subsec:Ex2} 

\begin{figure}
\centering
\includegraphics[scale=0.5]
{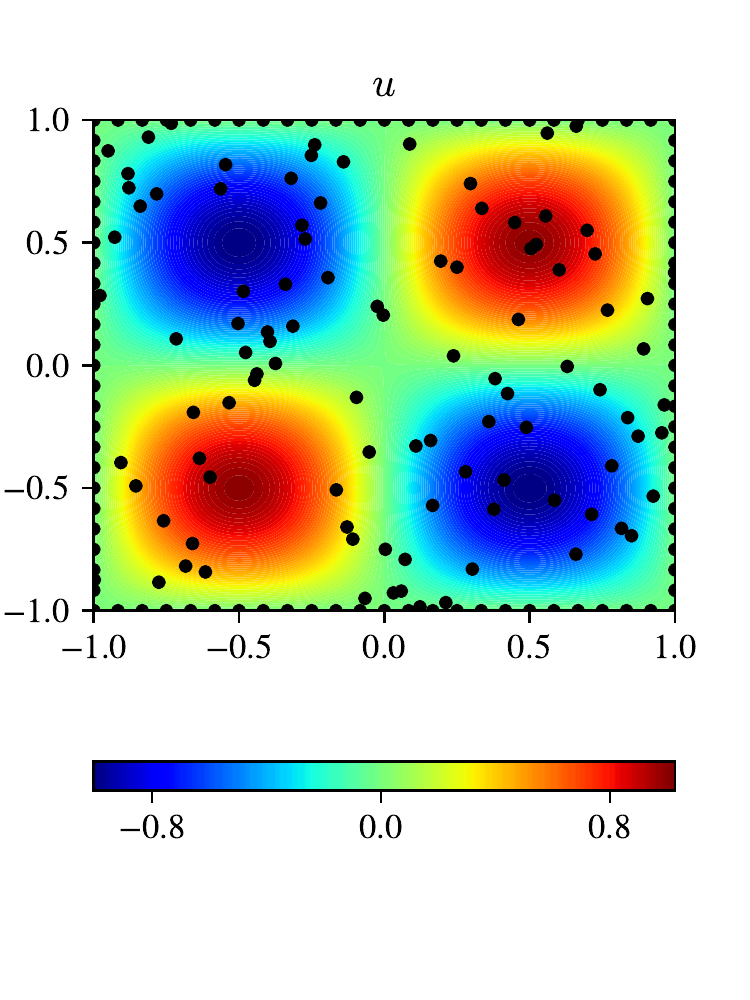}
\vspace{-1cm}
\caption{Example \ref{subsec:Ex2}: measurements of forward solution $\mathbf{u}$ and boundary measurements $\mathbf{b}$ with solution $u(x,y)=\sin(\pi x)\sin(\pi y)$.}
\label{fig:Ex_2_truth}
\end{figure}

\begin{table}
\centering
\begin{tabular}{c|c|c}
\hline
& &  $k \text{ (mean} \pm \text{std)}$  \\
\hline 
\multirow{ 2}{*}{$\sigma_u=0.01$}& EKI &  $0.999 \pm 0.006$  \\
&HMC &  $0.996 \pm 0.006$  \\
\hline
\multirow{ 2}{*}{$\sigma_u=0.1$}& EKI &  $0.988 \pm 0.017$  \\
&HMC &  $1.023 \pm 0.032$  \\ 
\hline
\end{tabular}
\caption{Example \ref{subsec:Ex2}: sample mean and standard deviation of parameter $k$ for EKI B-PINN and HMC B-PINN for $\sigma_u=0.01, 0.1$ noise levels. The true value of $k$ is 1.
\label{tab:ex2_confidence} } 
\end{table}

Next, we examine the 2D nonlinear diffusion-reaction equation in \cite{Yang_2021} as follows:
\begin{align}
\lambda\Delta u + ku^2 &= f, \quad (x,y)\in[-1,1]^2,\label{al:ex2}\\
u(x,-1) &= u(x,1) = 0,\\
u(-1,y) &= u(1,y) = 0,
\end{align}
where $\lambda = 0.01$ is known. Here, $k = 1$ is an unknown physical parameter, and the ground truth solution $u(x,y)=\sin(\pi x)\sin(\pi y)$. We construct the source term $f$ to satisfy the PDE with the given solution. For this problem, we have $N_u=100$ measurements $\mathbf{u}$ and $N_f=100$ residual points $\mathbf{f}$ both sampled via Latin Hypercube sampling over the spatial domain. Additionally, we have $N_b=100$ boundary points $\mathbf{b}$, which are equally spaced over the boundary. As in the previous example, we aim to estimate $k$ and $u$ with uncertainty estimates. The solution $u$ and measurements $\mathbf{u}$ can be seen in Figure \ref{fig:Ex_2_truth}.

\begin{table}
\centering
\begin{tabular}{c|c|cc|c|c}
\hline
& &  $e_{u}$  &  $e_{k}$ & Walltime & Iterations\\
\hline 
\multirow{ 2}{*}{$\sigma_u=0.01$}& EKI & 1.12\% & 0.04\%&2.47 seconds& 53\\
&HMC & 1.06\% & 0.03\% &63.26 seconds&-\\
\hline
\multirow{ 2}{*}{$\sigma_u=0.1$}& EKI & 3.64\% & 1.46\% &3.41 seconds& 66\\
&HMC & 2.53\% & 3.73\% &64.71 seconds&-\\ 
\hline
\end{tabular}
\caption{Example \ref{subsec:Ex2}: relative errors $e_u$ of the forward solution $u$ and $e_k$ of parameter $k$ for the noise levels $\sigma_u=0.01,0.1$, as well as average walltime. Average EKI iterations for the EKI B-PINN are also reported.
\label{tab:ex2_error} } 
\end{table}

\begin{figure}
\includegraphics[width=1\linewidth]{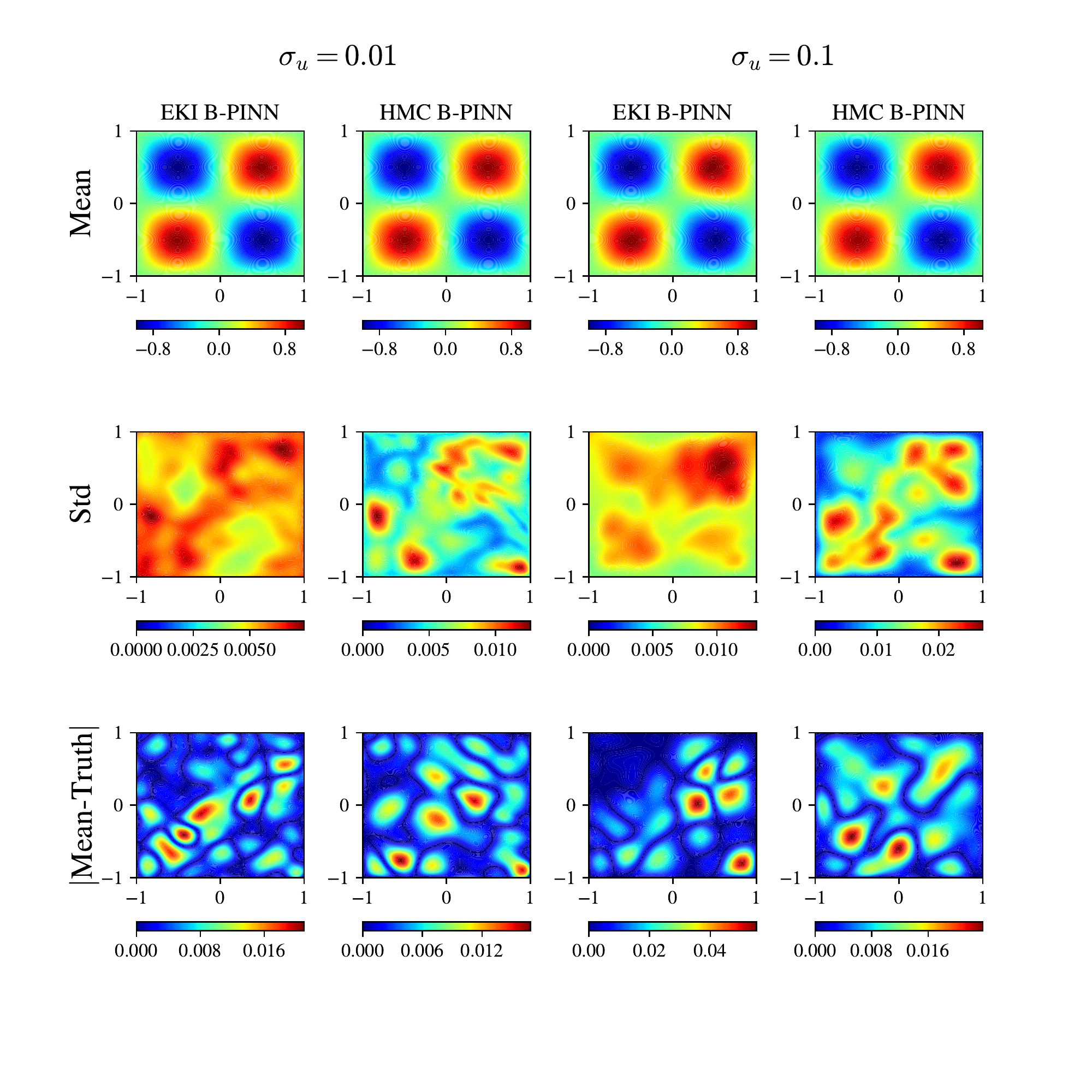}
\vspace{-2.0cm}
\caption{Example \ref{subsec:Ex2}: (Top Row) the sample mean of forward surrogate for EKI and HMC B-PINNs for different noise levels $\sigma_u$. (Middle Row) The standard deviation of the forward surrogate based on EKI and HMC B-PINNs for different noise levels. (Bottom Row) The absolute difference between the ground truth solution $u(x,y)=\sin(\pi y)\sin(\pi y)$ and the sample mean of the forward approximation for different noise levels.}
\label{fig:Ex_2_u}
\end{figure}

Table \ref{tab:ex2_confidence} shows the one standard deviation confidence interval of the B-PINN estimates for parameter $k$. Notably, the ground truth $k=1.0$ falls within one standard deviation of the mean estimates for both noise levels. Additionally, Figure \ref{fig:Ex_2_u} shows the sample mean, one standard deviation, and the error of the forward surrogate based on both approaches. Both B-PINNs provide reasonably good mean estimates of the true solution for two measurement noise levels. Moreover, the standard deviation for both B-PINNs increases as the measurement noise level increases as expected, indicating that the uncertainty estimates  provided are plausible. Although the standard deviations by the EKI B-PINN somewhat differ from those of the HMC B-PINN, both methods appear to agree on the rough locations of regions with higher uncertainty. For instance, when $\sigma_u=0.01$, For example, when $\sigma_u=0.01$, both methods exhibit large peaks around $(x,y)=(-1,0),(-0.3,0.7),(0.9,0.9),(1,-1)$. Similarly, major peaks can be seen around  at $(x,y)=(0.8,0.8),(-0.8,-0.8),(-0.8,0.8),(0.8,-0.8)$ for both methods when $\sigma_u=0.1$.

The relative error of the mean estimates of $u$ and $k$ and walltime for the B-PINN methods are presented in Table \ref{tab:ex2_error}. The mean approximations of $k$ for both B-PINNs show approximation errors around $1\%$ and $5\%$ of the ground truth for $\sigma_u=0.01$ and $\sigma_u=0.1$, respectively. Similarly, the sample mean approximation of the forward surrogate $u$ achieved a relative error less than $2\%$ and $5\%$, respectively. The EKI B-PINN approximates the forward solution and physical parameter reasonably well, with mean estimates comparable to the HMC. Nonetheless, the EKI B-PINN provides inference approximately 18 times faster than the HMC B-PINN.

\subsection{Kraichnan-Orszag system}
\label{subsec:Ex5}
We next consider the Kraichnan-Orszag model \cite{zou2022neuraluq} consisting of three coupled nonlinear ODEs describing the temporal evolution of a system composed of several interacting inviscid shear waves:
\begin{align}
\frac{du_1}{dt}& -a u_2 u_3=0, \label{al:ex5}\\
\frac{du_2}{dt}& - b u_1 u_3=0,\\ 
\frac{du_3}{dt}& +(a + b) u_1 u_2=0,\\
u_1(0)&=1.0,u_2=0.8, u_3(0)=0.5,
\end{align}
where $u_1$, $u_2$, and $u_3$ are solutions to the above system of ODEs and $a$ and $b$ are unknown physical parameters. We choose $a=b=1$ in this example. We place 12 equally spaced data points over $t\in[1,10]$, and observe $u_1$ and $u_3$ at each of these locations, and $u_2$ at the first 7 locations, and thus, $N_u=31$ for this example. We also utilize $N_f=300$ residual points (with $100$ points equally spaced over $t\in[0,10]$ for each equation in the system of ODEs). Initial conditions for the ODE are assumed to be unknown, thus $N_b=0$. As in \cite{zou2022neuraluq}, we place a Gaussian prior on $a$ and $b$ such that $a\sim\mathcal{N}(0,2)$ and $b\sim\mathcal{N}(0,2)$. Our goal is to estimate $u_1$, $u_2$, $u_3$, and parameters $a$ and $b$ with uncertainty estimates.

\begin{figure}
\includegraphics[width=1\linewidth]{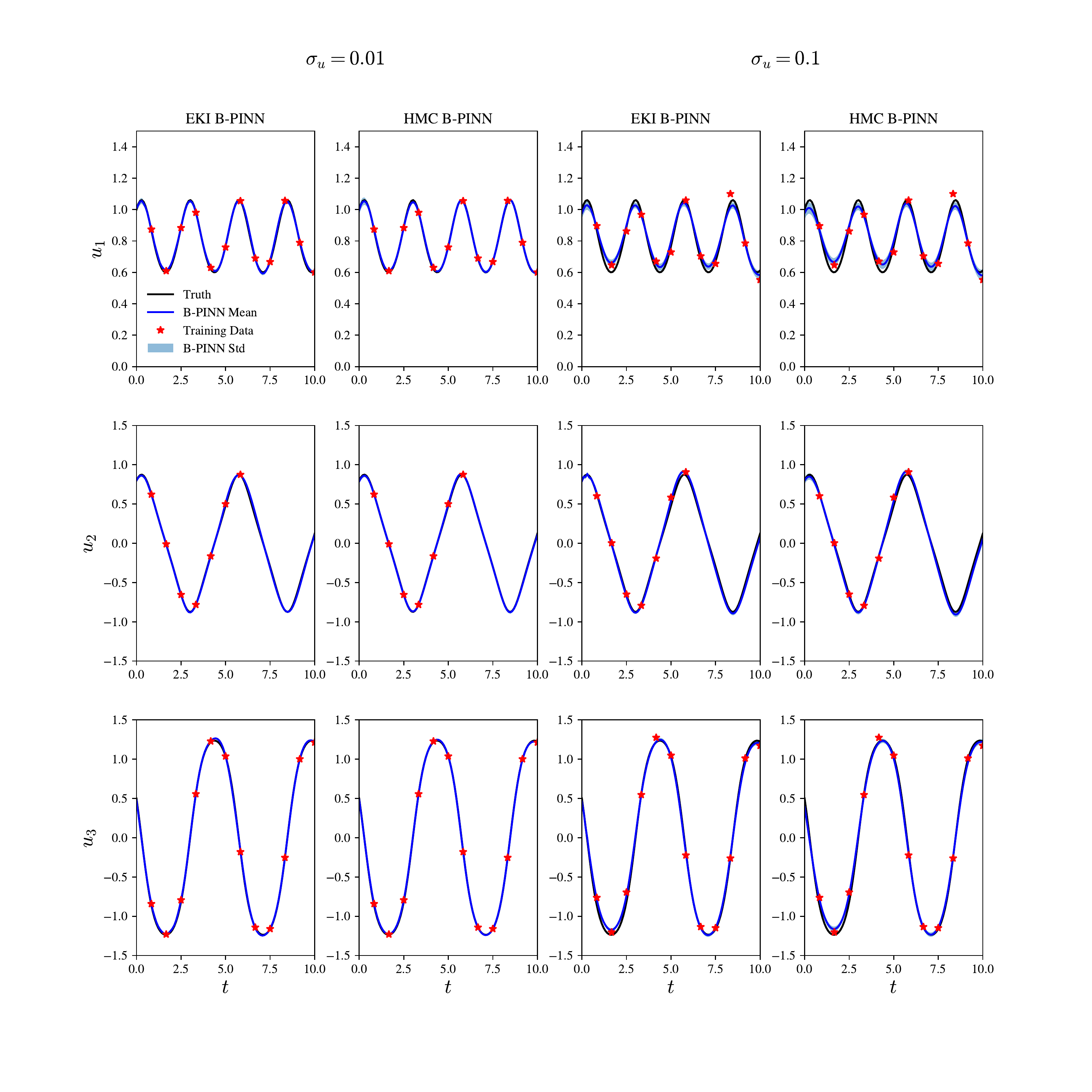}
\caption{Example \ref{subsec:Ex5}: the ground truth, sample mean, standard deviation, and measurements $\mathbf{u}$ for $u_1$ (top row), $u_2$ (middle row), and $u_3$ (bottom row) for the EKI B-PINN and HMC B-PINN with noise levels $\sigma_u=0.01, 0.1$.\label{fig:Ex_5_u}}
\end{figure}

Table \ref{tab:Ex_5_confidence} shows the mean and one standard deviation of the estimated $a$ and $b$. For the small noise level $\sigma_u=0.01$, the parameter estimates capture the truth $b$ within one standard deviation and $a$ within two standard deviations of the mean. For $\sigma_u=0.1$ noise level, both estimates find $b$ within two standard deviations of the mean. However,
$a$ is less well approximated for  both B-PINNs due to bias in the mean approximation. Furthermore, Figure \ref{fig:Ex_5_u} presents the approximation of the forward solutions ($u_1, u_2, u_3$) with the means and one standard deviation bands. For $\sigma=0.01$, the forward predictions closely match the true solutions with narrow uncertainty bands for the B-PINNs. For the case $\sigma=0.1$, the deviations between the mean prediction of $u_1$ and the ground truth are more pronounced, particularly at locations where the standard deviation is also larger, suggesting both B-PINNs offer informative uncertainty estimates.

Table \ref{tab:Ex_5_error} presents the mean relative errors for parameter estimations and forward approximations. The mean approximations of the EKI B-PINN are accurate and comparable to those of the HMC B-PINN for both noise levels, with the exception of parameter $a$ for $\sigma_u=0.1$, where both B-PINNs show less accuracy. However, the EKI method is approximately 25 times faster than the HMC method in this example.

\begin{table}
\centering
\begin{tabular}{c|c|cc}
\hline
& &  $a \text{ (mean} \pm \text{std)}$  &  $b \text{ (mean} \pm \text{std)}$  \\
\hline 
\multirow{ 2}{*}{$\sigma_u=0.01$}& EKI &  $0.953 \pm 0.023$  &  $1.003 \pm 0.018$  \\
&HMC &  $0.978 \pm 0.025$  &  $1.001 \pm 0.016$  \\
\hline
\multirow{ 2}{*}{$\sigma_u=0.1$}& EKI &  $0.847 \pm 0.033$  &  $1.027 \pm 0.026$  \\
&HMC &  $0.826 \pm 0.053$  &  $1.029 \pm 0.032$  \\ 
\hline
\end{tabular}
\caption{Example \ref{subsec:Ex5}: sample mean and standard deviation of parameters $a$ and $b$ for EKI B-PINN and HMC B-PINN for $\sigma_u=0.01, 0.1$ noise levels. The true values of $a=1, b = 1$. \label{tab:Ex_5_confidence}}
\end{table}

\begin{table}
\resizebox{\textwidth}{!}{%
\begin{tabular}{c|c|ccccc|c|c}
\hline
& &  $e_{u_1}$  &  $e_{u_2}$  &  $e_{u_3}$  &  $e_{a}$  &  $e_{b}$ & Walltime & Iterations \\
\hline 
\multirow{ 2}{*}{$\sigma_u=0.01$}& EKI & 1.04\% & 2.96\% & 1.63\% & 4.67\% & 0.35\%&2.93 seconds& 71 \\
&HMC & 0.66\% & 1.45\% & 1.06\% & 2.25\% & 0.14\% &93.34 seconds&- \\
\hline
\multirow{ 2}{*}{$\sigma_u=0.1$}& EKI & 3.49\% & 5.57\% & 4.45\% & 15.26\% & 2.73\% & 3.51 seconds &85\\
&HMC & 3.96\% & 6.49\% & 5.19\% & 17.42\% & 2.88\%&94.31 seconds &- \\ 
\hline
\end{tabular}
}
\caption{Example \ref{subsec:Ex5}: relative errors $e_u$ of the forward solution $u$, $e_a$ of parameter $a$, and $e_b$ of parameter $b$ for the noise levels $\sigma_u=0.01, 0.1$, as well as average walltime. Average EKI iterations are also reported. \label{tab:Ex_5_error}}
\end{table}

\subsection{Burgers' Equation}
\label{subsec:Ex6}
\begin{figure}
\centering
\includegraphics[scale=0.5]{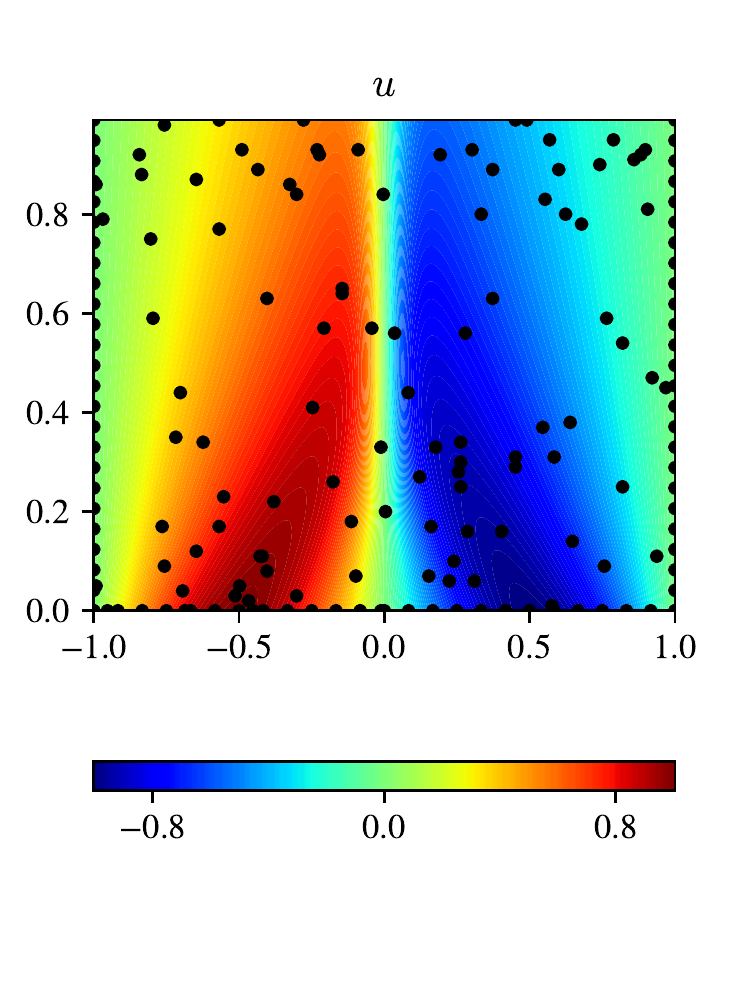}
\vspace{-1.0cm}
\caption{Example \ref{subsec:Ex6}: measurements of forward solution $\mathbf{u}$ and boundary measurements $\mathbf{b}$ with solution $u(x,t)$ over $[-1,1]\times[0,\frac{3}{\pi}]$.}
\label{fig:Ex_6_truth}
\end{figure}
We now consider the following 1D time-dependent Burgers' equation motivated by \cite{RAISSI2019}:
\begin{align}
u_{t} + u u_{x} - \nu u_{xx}&=0,\quad\quad\quad\hspace{2.2mm}\quad x\in[-1,1], \label{al:ex6}\\
u(x,0) &= -\sin(\pi x),\quad x\in[-1,1],\\
u(-1,t) &= u(1,t) = 0,
\end{align}
where $\nu=\frac{0.1}{\pi}$ is an unknown physical parameter, and $t\in[0,\frac{3}{\pi}]$. A reference solution $u$ for \eqref{al:ex6} is found in \cite{BASDEVANT198623}, where we evaluate on a $256\times100$ equally spaced grid over the domain. We randomly sampled $N_u=100$ measurements $\mathbf{u}$ from the solution grid and placed $N_b=75$ equally spaced boundary points over the boundary, which is visualized with the solution in Figure \ref{fig:Ex_6_truth}. Residual points $\mathbf{f}$ were sampled using Latin hypercube sampling over $(x,t)\in[-1,1]\times[0,\frac{3}{\pi}]$ with $N_f=100$. Instead of directly approximating $\nu$, we approximate the transformed parameter $\log \nu$ and place the prior $\log \nu\sim\mathcal{N}(0,3)$. Additionally, for this example, we choose the standard deviation of the likelihood for the residual $\sigma_f=0.1$. In this example, we aim to infer $\nu$ and $u$ with corresponding uncertainty estimates.

\begin{table}
\centering
\begin{tabular}{c|c|c}
\hline
& &  $\nu$ $(\times 10^3) \text{ (mean} \pm \text{std)}$  \\
\hline 
\multirow{ 2}{*}{$\sigma_u=0.01$}& EKI &  $32.449 \pm 0.602$  \\
&HMC &  $30.957 \pm 1.027$  \\
\hline
\multirow{ 2}{*}{$\sigma_u=0.1$}& EKI &  $33.740 \pm 0.808$  \\
&HMC &  $34.174 \pm 3.523$  \\ 
\hline
\end{tabular}
\caption{Example \ref{subsec:Ex6}: the sample mean and standard deviation of parameter $\nu$ for EKI B-PINN and HMC B-PINN for $\sigma_u=0.01, 0.1$ noise levels. The true value $\nu= \frac{0.1}{\pi}\approx 0.031831$.}. 
\label{tab:ex6_confidence}
\end{table}

\begin{figure}
\includegraphics[width=1\linewidth]{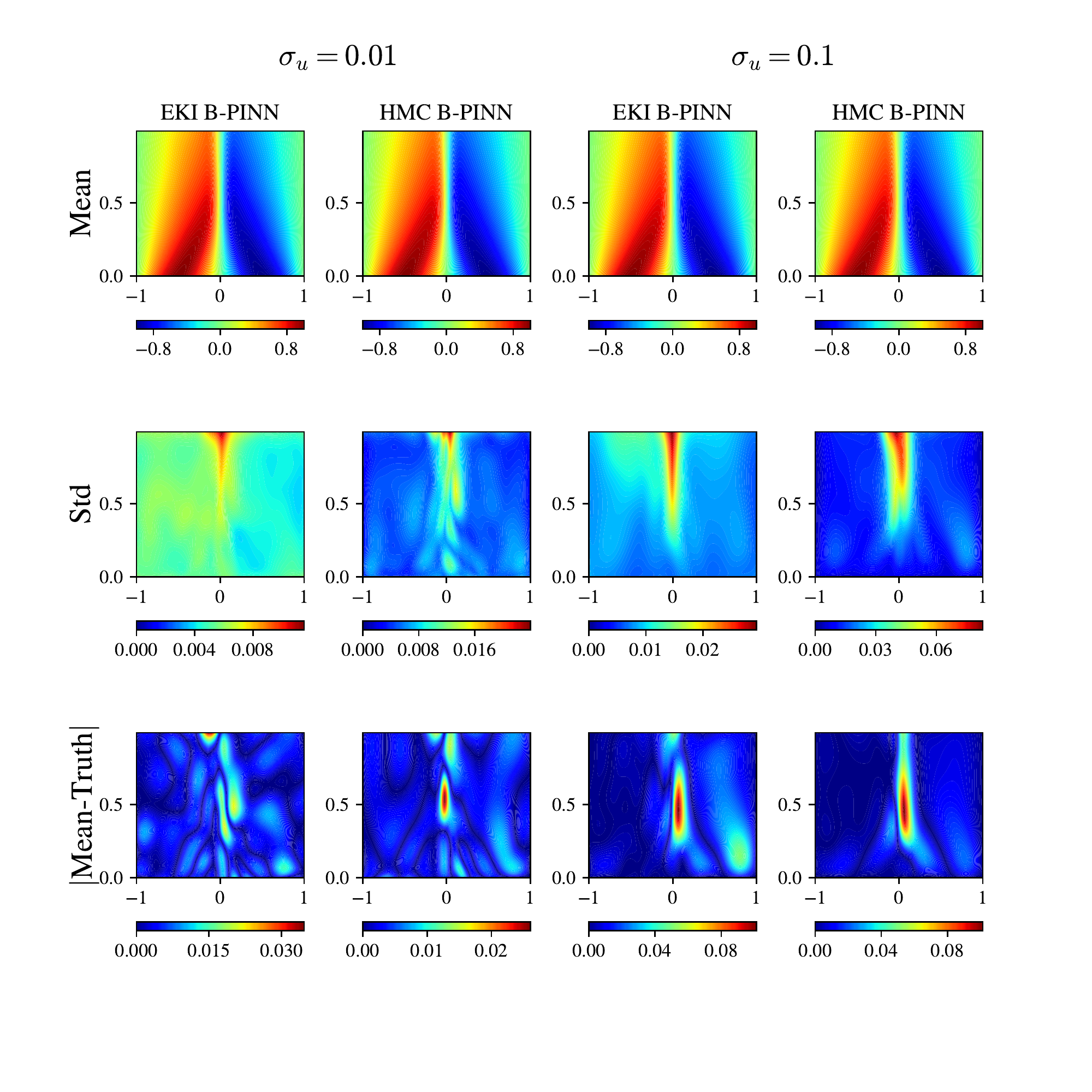}
\vspace{-2cm}
\caption{Example \ref{subsec:Ex6}: (Top Row) the sample mean of forward surrogate for EKI and HMC B-PINNs for different noise levels $\sigma_u$. (Middle Row) The standard deviation of the forward surrogate based on EKI and HMC B-PINNs for different noise levels. (Bottom Row) The absolute difference between the ground truth solution $u(x,y)$ and the sample mean of the forward approximation for different noise levels.}
\label{fig:Ex_6_u}
\end{figure}

\begin{table}
\centering
\begin{tabular}{c|c|cc|c|c}
\hline
& &  $e_{u}$  &  $e_{\nu}$  & Walltime & Iterations \\
\hline 
\multirow{ 2}{*}{$\sigma_u=0.01$}&  EKI & 1.04\% & 1.94\%& 5.90 seconds& 71\\
&HMC & 0.77\% & 2.75\%  &62.57 seconds&-\\
\hline
\multirow{ 2}{*}{$\sigma_u=0.1$}& EKI & 3.48\% & 6.00\%& 6.72 seconds& 116\\
&HMC & 3.07\% & 7.36\% &62.92 seconds&-\\ 
\hline
\end{tabular}
\caption{Example \ref{subsec:Ex6}: relative errors $e_u$ of the forward solution $u$ and $e_\nu$ of parameter $\nu$ for the noise levels $\sigma_u=0.01,0.1$, as well as the average walltime. Average iterations for EKI B-PINN are also reported. 
\label{tab:ex6_error}} 
\end{table} 

Table \ref{tab:ex6_confidence} shows the posterior mean and one standard deviation of the estimated parameter $\nu$. Both methods provide accurate mean approximations of the parameter, and the true $\nu$ is contained within the two standard deviation bounds for $\sigma_u=0.01$, and lies just outside the one standard deviation confidence interval from the EKI B-PINN. For $\sigma_u=0.1$, EKI B-PINN does provide a slightly better mean estimate, but it underestimates the uncertainty compared to the HMC.

Figure \ref{fig:Ex_6_u} displays the posterior mean and standard deviation of $u$ estimated by both methods, as well as their corresponding approximation error. As expected, the uncertainty in both B-PINNs increases with the measurement noise level. Notably, larger uncertainty is clustered around $x=0$, and greater errors are observed in the same region in the error plot, which suggests that the uncertainty estimates are informative as an error indicator.

Table \ref{tab:ex6_error} provides mean relative error for $\nu$ and $u$. Both methods can achieve reasonably good accuracy with relative errors of less than $2\%$ for both $\nu$ and $u$ when $\sigma_u=0.01$ and $8\%$ for $\sigma=0.1$. The inference results from  EKI are obtained approximately 10 times faster than the HMC method.

\subsection{Diffusion-Reaction Equation with Source Localization}
\label{subsec:Ex8}

\begin{figure}
\centering
\includegraphics[scale=0.5]{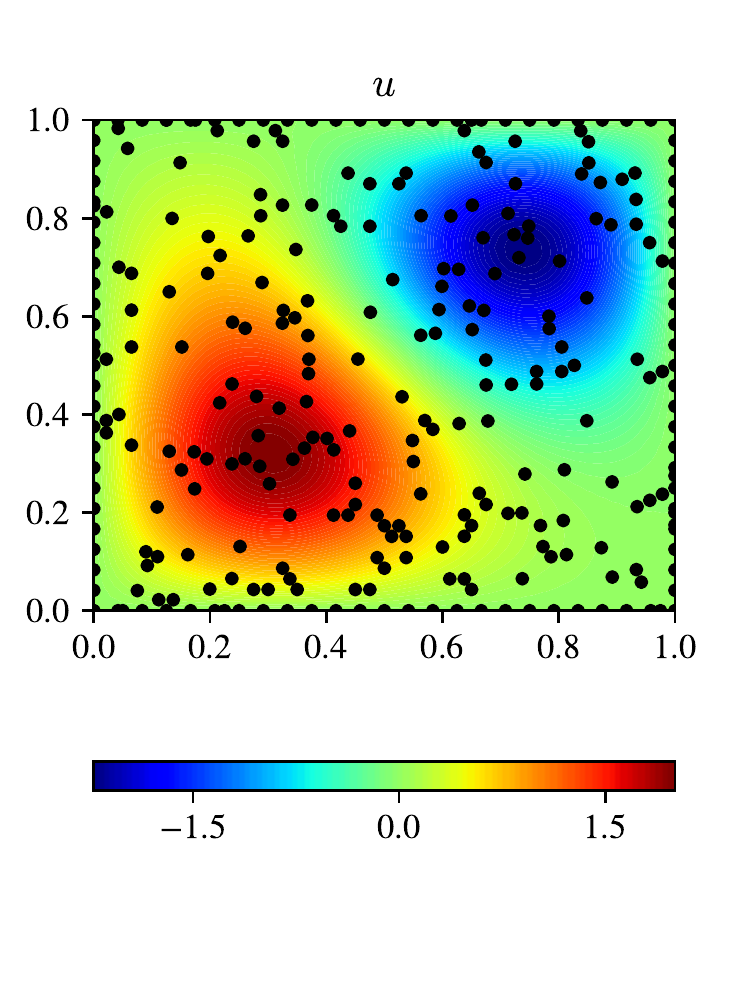}
\vspace{-1cm}
\caption{Example \ref{subsec:Ex8}: measurements of forward solution $\mathbf{u}$ and boundary measurements $\mathbf{b}$ with solution $u(x,t)$. \label{fig:Ex_8_truth}}
\end{figure}

Finally, we consider the following source inversion problem for a two-dimensional linear diffusion-reaction equation inspired by the example \cite{Yang_2021}:
\begin{align}
-\lambda \Delta u - f_2 &= f_1,\quad (x,y)\in[0,1]^2,\label{al:ex8}\\
u(0,y) = u(1,y) &= 0,\\
u(x,0) = u(x,1) &= 0,
\end{align}
where $\lambda=0.02$ is the known diffusivity and $f_1(x) = 0.1\sin(\pi x)\sin(\pi y)$ is a known forcing. The goal is to infer the location parameters of the field $f_2$, corresponding to three contaminant sources of the unknown location in the following equation:
 \begin{align}
 f_2(\mathbf{x}) = \sum_{i=1}^{3} k_i\exp\left(-\frac{||\mathbf{x}-\mathbf{x}_i||^2}{2\ell^2}\right).\label{al:forcing}
 \end{align}
 Here, $\mathbf{k} =(k_1, k_2, k_3) = (2,-3,0.5)$ and $\ell=0.15$ are known constants and parameters $\mathbf{x}_1 = (0.3,0.3)$, $\mathbf{x}_2 = (0.75,0.75)$, $\mathbf{x}_3 = (0.2,0.7)$ are parameters to be recovered. The prior distributions
on the locations $\mathbf{x}_i=(x_i,y_i)$ are chosen to be log-normal, that is $\log(x_i)\sim \mathcal{N}(0,1)$ and $\log(y_i)\sim \mathcal{N}(0,1)$ for $i=1,2,3$. The measurements $\mathbf{u}$ are generated by solving \eqref{al:ex8} using the \emph{solvepde} function in Matlab with 1893 triangle meshes and sampling randomly $N_u=100$ points from among the nodes and $N_b=100$ equally spaced points along the domain's boundary, shown in Figure \ref{fig:Ex_8_truth}. In addition, $N_f=100$ residual points were obtained via Latin Hypercube Sampling within the domain.

\begin{table}
\centering
\resizebox{\textwidth}{!}{%
\begin{tabular}{c|c|cccccc}
\hline
& &  $x_1 \text{ (mean} \pm \text{std)}$  &  $y_1 \text{ (mean} \pm \text{std)}$  &  $x_2 \text{ (mean} \pm \text{std)}$  &  $y_2 \text{ (mean} \pm \text{std)}$  &  $x_3 \text{ (mean} \pm \text{std)}$  &  $y_3 \text{ (mean} \pm \text{std)}$  \\
\hline 
\multirow{ 2}{*}{$\sigma_u=0.01$}& EKI &  $0.299 \pm 0.006$  &  $0.291 \pm 0.006$  &  $0.754 \pm 0.004$  &  $0.751 \pm 0.004$  &  $0.201 \pm 0.019$  &  $0.690 \pm 0.019$   \\
&HMC &  $0.294 \pm 0.003$  &  $0.300 \pm 0.005$  &  $0.749 \pm 0.004$  &  $0.747 \pm 0.003$  &  $0.198 \pm 0.015$  &  $0.688 \pm 0.010$ \\
\hline
\multirow{ 2}{*}{$\sigma_u=0.1$}& EKI &  $0.300 \pm 0.006$  &  $0.297 \pm 0.006$  &  $0.754 \pm 0.004$  &  $0.748 \pm 0.005$  &  $0.196 \pm 0.020$  &  $0.709 \pm 0.021$  \\
&HMC &   $0.299 \pm 0.005$  &  $0.300 \pm 0.004$  &  $0.748 \pm 0.005$  &  $0.750 \pm 0.003$  &  $0.210 \pm 0.017$  &  $0.682 \pm 0.015$   \\ 
\hline
\end{tabular}}
\caption{Example \ref{subsec:Ex8}: sample mean and standard deviation of parameters $\mathbf{x}_1=(x_1,y_1)=(0.3,0.3)$, $\mathbf{x}_2=(x_2,y_2)=(0.75,0.75)$, and $\mathbf{x}_3=(x_3,y_3)=(0.2,0.7)$ for EKI B-PINN and HMC B-PINN at $\sigma_u=0.01,0.1$ noise levels.\label{tab:ex8_confidence}}
\end{table}

\begin{figure}
\includegraphics[width=1\linewidth]{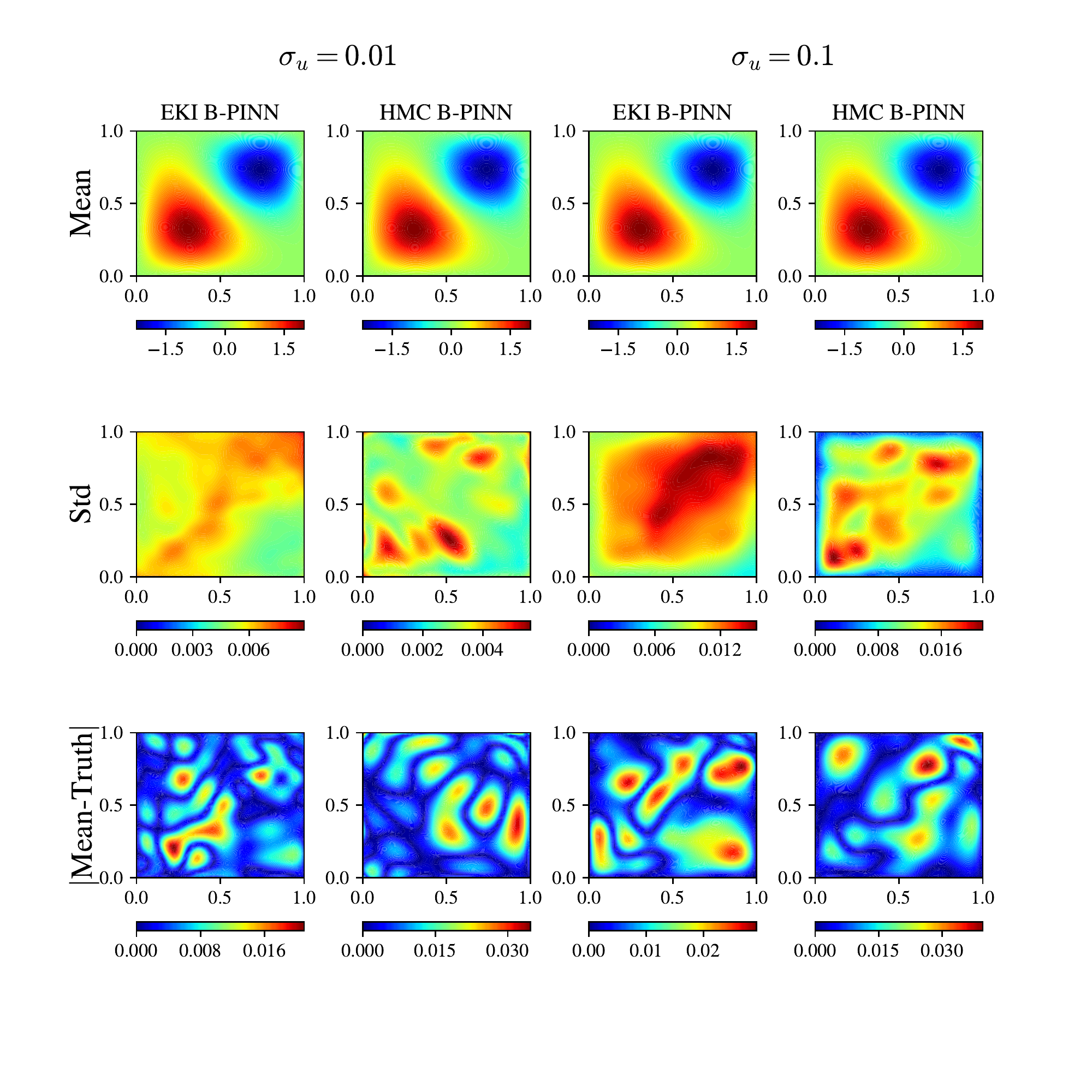}
\vspace{-2cm}
\caption{Example \ref{subsec:Ex8}: (Top Row) the sample mean of forward surrogate for EKI and HMC B-PINNs for different noise levels $\sigma_u$. (Middle Row) The standard deviation of the forward surrogate based on EKI and HMC B-PINNs for different noise levels. (Bottom Row) The absolute difference between the ground truth solution $u(x,y)$ and the sample mean of the forward approximation for different noise levels. \label{fig:Ex_8_u}}
\end{figure}

The posterior mean and standard deviation of the locations $\mathbf{x}_1$, $\mathbf{x}_2$ and $\mathbf{x}_3$ in Table \ref{tab:ex8_confidence} show an accurate estimation of most of the centers of contaminants, with all values falling within contained within two standard deviations and most within one standard deviation of the mean estimates. The corresponding solution obtained for both B-PINNs, with the first two moments and approximation error shown in Figure \ref{fig:Ex_8_u}, shows high accuracy regarding the mean prediction. Although the uncertainty estimates for the surrogate solutions by both B-PINNs somewhat differ for both noise levels, their magnitudes are comparable over the domain for both noise levels. Furthermore, when $\sigma_u=0.01$, both methods identify some similar bulk regions with higher uncertainty, particularly near $(x,y)=(0.5,0.25)$ and $(0.75,0.75)$. Similarly for  $\sigma_u=0.1$, the regions of higher uncertainty show rough similarities between the two methods, with larger uncertainties clustering near the right diagonal of the domain.

Table \ref{tab:ex8_error} compares the mean relative errors of physical parameters $\mathbf{x}_i$ and forward solution $u$ for two noise levels, as well as the walltime and number of EKI iterations. Consistent with the previous examples, EKI-based inference is approximately 25-fold faster than that of the HMC.

\begin{table}
\centering
\resizebox{\textwidth}{!}{%
\begin{tabular}{c|c|ccccccc|c|c}
\hline
& &  $e_{u}$  &  $e_{x_1}$  &  $e_{y_1}$  &  $e_{x_2}$  &  $e_{y_2}$  &  $e_{x_3}$  &  $e_{y_3}$ & Walltime & Iterations \\
\hline 
\multirow{ 2}{*}{$\sigma_u=0.01$}& EKI & 0.72\% & 0.28\% & 2.88\% & 0.60\% & 0.11\% & 0.73\% & 1.46\%  &2.66 seconds & 59\\
&HMC & 0.72\% & 0.28\% & 2.88\% & 0.60\% & 0.11\% & 0.73\% & 1.46\%  &68.50 seconds&-\\
\hline
\multirow{ 2}{*}{$\sigma_u=0.1$}& EKI & 1.30\% & 0.08\% & 1.02\% & 0.53\% & 0.30\% & 2.11\% & 1.24\% &2.97 seconds &67\\
&HMC & 1.43\% & 0.45\% & 0.17\% & 0.23\% & 0.06\% & 5.08\% & 2.62\% &68.58  seconds&-\\ 
\hline
\end{tabular}
}
\caption{Example \ref{subsec:Ex8}: relative errors $e_u$ of the forward solution $u$ and $e_{x_i}$, $e_{y_i}$ for source locations $x_i$ and $y_i$ for $i=1,2,3$ respectively for the noise levels $\sigma_u=0.01,0.1$, as well as average walltime. Average EKI iterations for the EKI B-PINN are also reported.
\label{tab:ex8_error}} 
\end{table}
\section{Summary}
\label{sec:summary}

This paper presents a new and efficient inference method for B-PINNs, utilizing Ensemble Kalman Inversion (EKI) to infer physical and neural network parameters. We demonstrate the applicability and performance of our proposed approach using several benchmark problems. Interestingly, while EKI methodology theoretically only provides unbiased inference with Gaussian priors and linear operators, our results show that it can still provide reasonable inference in non-linear and non-Gaussian settings, which is consistent with findings in other literature \cite{Botha2022,DUFFIELD2022109523}. In all examples, our proposed method delivers comparable inference results to HMC-based B-PINNs, but with around 8-30 times speedup. Furthermore, EKI B-PINNs can provide informative uncertainty estimates for physical model parameters and forward predictions at different noise levels comparable to the results of HMC B-PINNs. In cases where more detailed uncertainty quantification is necessary, our proposed approach can serve as a good initialization for other inference algorithms, such as HMC, to produce better results with reduced computational costs.

Besides, it is worth noting that our study did not investigate the case of a large dataset or residual points.  In such cases, naive approaches to EKI would be computationally expensive due to the cubic scaling of EKI with the size of the dataset. In such a case,
mini-batching techniques proposed in \cite{Kovachki_2019} for EKI to train NNs with large datasets, may help to overcome this challenge. Finally, we acknowledge that the EKI requires storing an ensemble of $J$ neural network parameter sets, which can be memory-demanding for large ensemble sizes and large networks. In this situation, Dimension reduction techniques may help address this issue. We plan to investigate this strategy in future work.
\clearpage

\bibliographystyle{plain}
\bibliography{biblio}

\end{document}